\PassOptionsToPackage{table,dvipsnames}{xcolor}
\documentclass{muxlab}
\usepackage{xspace}
\usepackage{amsfonts}
\usepackage{mathtools}
\usepackage{algorithmicx}
\usepackage{algpseudocode}
\usepackage{algorithm}
\usepackage{wrapfig}
\usepackage{needspace}
\usepackage{makecell}

\newcommand{\fitwidth}[1]{%
  \resizebox{\ifdim\width>\linewidth\linewidth\else\width\fi}{!}{#1}}

\DeclareMathOperator*{\argmax}{\mathsf{argmax}}

\definecolor{LoopedFlowPurple}{HTML}{5B2D8E}
\definecolor{LoopedFlowLight}{HTML}{EDE7F6}
\definecolor{LoopedFlowDeep}{HTML}{311B52}
\definecolor{OursBg}{HTML}{EDE7F6}
\definecolor{ours}{HTML}{6B4FBB}
\makeatletter
\let\LF@fs@ruled\fs@ruled
\def\fs@ruled{\LF@fs@ruled\def\@fs@pre{\normalcolor\hrule height.8pt depth0pt \kern2pt}}
\makeatother
\algrenewcommand\algorithmicrequire{\textbf{Input:}}
\algrenewcommand\algorithmicensure{\textbf{Output:}}
\algrenewcommand\algorithmiccomment[1]{\hfill{\color{black!55}\scriptsize$\triangleright$~#1}}
\newcommand{\algreturn}{\textbf{return}\ }

\usepackage{cleveref}
\usepackage{backref}

\crefname{equation}{}{}
\crefname{appendix}{Appendix}{Appendices}
\Crefname{appendix}{Appendix}{Appendices}

\renewcommand*{\backref}[1]{}
\renewcommand*{\backrefalt}[4]{\ifcase#1\relax\or(page #2)\else(pages #2)\fi}

\newcommand{\ours}{Floodgates\xspace}
\newcommand{\tauacc}{\tau}

\title{Looped Models are Fast Multi-Token Predictors}

\renewcommand{\authorlist}{%
  {\sffamily\color{muxfg}%
    \textbf{Ayhan Suleymanzade}\textsuperscript{1},\;\;
    \textbf{Sean McLeish}\textsuperscript{1,2},\;\;
    \textbf{Sehoon Kim}\textsuperscript{3},\;\;
    \textbf{Jinwoo Kim}\textsuperscript{3},\;\;
    \textbf{Antoine Bosselut}\textsuperscript{1}%
  }%
}
\renewcommand{\affiliationlist}{%
  {\small\textsuperscript{1}EPFL\quad \textsuperscript{2}University of Maryland\quad \textsuperscript{3}KAIST}%
}
\abstract{
Looped language models can adapt computation for predicting a token without a proportional rescaling in unique model parameters by adjusting the number of loops. This property has been used for self-speculative decoding, taking shallow loops as fast drafters and deep loops as verifiers, which is effective since drafting still uses full parameters. In this paper, we show that shallow loops can be also predictive of several future tokens and leverage this to drastically accelerate the inference of looped language models. To that end, we present \textbf{\ours}, a simple self-speculative decoding method that appends a block of mask tokens as a suffix and learns via low-rank adaptation to draft these tokens in parallel. Across seven models, five benchmarks, a range of batch sizes, and both greedy and sampled decoding, \ours is the fastest lossless decoding method in most cases, attaining up to 7.5$\times$ acceleration. Through analysis, we show that this speedup is enabled by all-depth training of looped models, with the shallow drafter's ability to leverage full-depth computation by conditioning on deep prefix cache evaluated during verification.
\\[2pt]
\textbf{Code:} \url{https://github.com/MisakiTaro0414/floodgates}%
}

\renewcommand{\topfraction}{0.9}\renewcommand{\bottomfraction}{0.6}\renewcommand{\textfraction}{0.07}\renewcommand{\floatpagefraction}{0.7}
\tcbset{after skip=0pt}
\begin{document}
\maketitle

\titlespacing*{\section}{0pt}{1ex plus .5ex minus .2ex}{2.3ex plus .2ex}
\vspace{-0.1em}
\section{Introduction}
\label{sec:intro}
\vspace{-0.1em}
\titlespacing*{\section}{0pt}{3.5ex plus 1ex minus .2ex}{2.3ex plus .2ex}

\begin{wrapfigure}[17]{r}{0.45\linewidth}
\vspace{-1.5em}
\centering
\includegraphics[width=\linewidth]{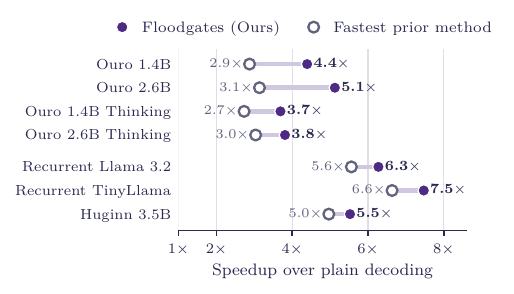}
\vspace{-2em}
\caption{\textbf{\ours is up to 7.5$\bm{\times}$ faster than autoregressive generation and faster than prior methods on every model.} Speedup over plain decoding under greedy decoding at batch size one, averaged over the benchmarks of each model. Open circles show the fastest prior lossless method on each model.}
\label{fig:teaser}
\end{wrapfigure}

Autoregressive (AR) transformer language models produce text one token after another. \begingroup\renewcommand{\thefootnote}{}\footnotetext{Correspondence to: \href{mailto:ayhan.suleymanzade@epfl.ch}{\texttt{ayhan.suleymanzade@epfl.ch}}.}\endgroup In each layer, the hidden state of the new token is updated by attending to the hidden states of all earlier tokens by the preceding layer. A token can therefore be generated only after all of the tokens before it, leading to a serialized bottleneck in speed~\citep{santilli2023accelerating,suleymanzade2026mux}. Due to local regularities in natural language~\citep{church-hanks-1990-word}, many upcoming tokens can be easy to predict, and speculative decoding uses this to accelerate  generation~\citep{leviathan2023fast,chen2023accelerating}. A fast drafter proposes several tokens, and the model verifies all of them in one forward pass. Yet, the drafter is usually a small separate network~\citep{li2024eagle,chen2026dflash} or a subset of the layers~\citep{zhang2024draft,elhoushi2024layerskip,liu2024kangaroo}, computed with only a fraction of the parameters of the model. Thus, the drafters are usually prevented from using the full knowledge encoded in all layers of the model, leading to a limited verification rate of their drafts.

Looped language models naturally avoid this limitation. They process each token through a group of layers repeatedly~\citep{dehghani2019universal,geiping2025scaling,zhu2025scaling}, so they can reduce their computation for faster generation by shallowing their loop without a proportional decrease in their parameters. This allows for an effective self-speculative decoding scheme that uses the deep loop to verify each token drafted by a shallow loop, in which the drafter now has access to the full knowledge within the model since every loop runs the same block. Yet, existing methods either draft a single token per loop~\citep{cho2026loopspec,ha2026wavefront} or attain good acceptance rate only when the model runs many loops~\citep{li2026shallow}.  Moreover, they often rely on \emph{pipelining} for parallelism, starting the next token before the current one has finished its loops. As tokens at different loops are batched together, the methods need to maintain separate, per-depth prefix cache across all batched loops, diminishing their utility when serving many requests at once where memory becomes a bottleneck.

In this paper, we show that a single loop of a looped language model can be adapted to draft several future tokens in parallel while attending to the final-loop prefix cache computed during verification, drastically accelerating inference. Our first intuition is that looped training can make shallow loops predictive of several future tokens, as it supervises every depth and hence pressures each future token to be predicted as early as possible, encouraging shallow hidden states of past tokens to be already helpful (\Cref{sec:dense}).
The second is that the shallow drafter can make use of full-depth computation through the prefix cache, which holds the keys and values of earlier tokens at the final loop. The model computes this cache during verification, so the drafter can attend to it ``for free''. We present \textbf{Floodgates}, a simple self-speculative decoding method that appends a block of mask tokens as a suffix and learns to read the final-loop prefix cache to draft these tokens in parallel using a single loop iteration. We use low-rank adapters~\citep{hu2022lora} that are active only while drafting, so the verifier parameters do not change and thus decoding remains lossless. On models that run many loops, where a single loop costs only a small fraction of a decoding step, \ours further refines this draft with additional passes of the pretrained loop (\emph{elastic drafting}) to balance drafting cost and acceptance without extra training. Across seven looped models and five math, coding, and reasoning benchmarks, \ours decodes faster than all prior lossless methods under greedy decoding and in most cases under sampling (\Cref{fig:teaser}). An illustration is given in \Cref{fig:overview}.
In summary, we make the following contributions:

\begin{figure}[t]
\centering
\includegraphics[width=\linewidth]{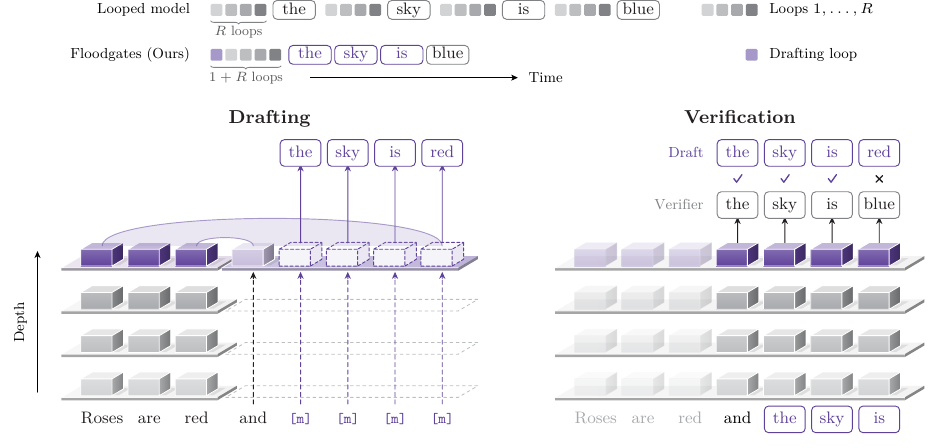}
\caption{\textbf{\ours drafts a block of tokens with a single loop and verifies it in one decoding step.} (Left) A single loop reads the deep prefix cache \textcolor[HTML]{5F429A}{(purple)} computed during the previous verification and drafts several tokens \textcolor[HTML]{A699CA}{(lilac)} in parallel from mask tokens. (Right) The full model verifies the drafts in one decoding step, accepts them up to the first mismatch, and produces the prefix cache for the next draft. (Top) This allows the shallow drafter to leverage full-depth computation for fast speculative decoding.
}
\label{fig:overview}
\end{figure}

\begin{enumerate}[label=(\arabic*),leftmargin=*,itemsep=1pt,topsep=2pt]
\item We introduce \ours, a simple self-speculative decoding method that adapts a single loop via low-rank adapters to draft several tokens in parallel from mask tokens while reading the deep prefix cache from verification, and refines the draft with elastic passes of the pretrained loop on deep models (\Cref{sec:method}). It overcomes the limitations of existing methods by effective multi-token prediction and its access to the deep prefix cache.

\item We evaluate \ours on seven looped models and five math, coding and reasoning benchmarks. \ours is the fastest lossless decoding method on every model and every task under greedy decoding and in most cases under sampling (\Cref{sec:main_results}). It decodes up to $5.1\times$ faster than plain decoding on average with a single loop on 4-loop models and up to $7.5\times$ faster with elastic drafting on 32-loop models, and on some models, it is nearly twice as fast as the best prior method. At the largest batch size that fits in memory, it reaches up to $1.9\times$ the throughput of the strongest baseline method (\Cref{sec:batch}).

\item Through controlled comparisons, we show that self-speculative drafting with a single loop iteration gives larger speedups than drafting with a separate smaller network (EAGLE, DFlash) or with the first layers of a non-looped model (Llama~3.2, LayerSkip), supporting the utility of drafting with all of the parameters of the model (\Cref{sec:dense}). We also show that deep prefix cache accelerates decoding up to $1.3\times$ (\Cref{sec:dense}), and that elastic drafting maximizes acceleration by optimizing the cost-acceptance tradeoff across verifiers of varying depths (\Cref{sec:controls}).
\end{enumerate}

\section{Background}
\label{sec:prelim}
Let $V$ denote a token vocabulary, and let $\Delta^{|V|-1}$ be the probability simplex over $V$, whose vectors ${\bf p}\in \Delta^{|V|-1}$ we index by ${\bf p}(v)\in\mathbb{R}$. An autoregressive language model generates text ${\bf x}=({\bf x}^1,{\bf x}^2,\dots)$ with ${\bf x}^t\in V$ sequentially, predicting each next token from a conditional distribution $p(\cdot\mid{\bf x}^{\le t})\in\Delta^{|V|-1}$. We write the cross-entropy between a predicted distribution ${\bf p}\in\Delta^{|V|-1}$ and a target token $v\in V$ as $\mathsf{CE}({\bf p},v)\coloneqq-\log{\bf p}(v)$.

\paragraph{Looped language models.}
Looped language models compute a hidden state ${\bf z}^t\in\mathbb{R}^d$ for each token by recurrently applying a shared block of layers $f_\theta$ \citep{dehghani2019universal, geiping2025scaling, zhu2025scaling}:
\begin{equation}
\label{eq:loop}
{\bf z}^t_r = f_\theta\bigl({\bf z}^t_{r-1};\,{\bf c}^{<t}_r\bigr), \qquad r=1,\dots,R,
\end{equation}
where ${\bf z}^t_0\in\mathbb{R}^d$ is computed from the token embedding ${\bf E}_{{\bf x}^t}\in\mathbb{R}^d$, and ${\bf c}^{<t}_r$ denotes the context at loop $r$, that is, the keys and values produced by earlier tokens in loop $r$. We refer to the context at the final loop, ${\bf c}^{<t}_R$, as the \emph{deep prefix cache}. The shared block updates the hidden state at each step, $f_\theta(\cdot\,;{\bf c}^{<t}_r):\mathbb{R}^d\to\mathbb{R}^d$, and an output head $g:\mathbb{R}^d\to\Delta^{|V|-1}$ predicts the next token from the final state, $p(\cdot\mid{\bf x}^{\le t})=g({\bf z}^t_R)$. We call each application of $f_\theta$ a loop, the $R$ loops that produce a token a decoding step, and $R$ the depth of the model.
In practice, looped models are trained with a next-token loss not only at the final loop $R$, but also at earlier loops, either by randomizing depth \citep{geiping2025scaling, mcleish2025teaching} or by supervising every loop \citep{zhu2025scaling}:
\begin{equation}
\label{eq:loop_loss}
\mathcal{L}_{\mathsf{loop}}(\theta)\coloneqq\mathbb{E}_{\bf x}\Bigl[\,\sum_{t}\sum_{r=1}^{R}\lambda_r\,\mathsf{CE}\bigl(g({\bf z}^t_r),\,{\bf x}^{t+1}\bigr)\Bigr],
\end{equation}
where $\lambda_r\ge0$ weights the loss at loop $r$. While (\ref{eq:loop_loss}) trains every loop to output the immediate next token, no loop is trained to output the tokens that follow it. An early loop can therefore directly read out the next token, but not several future tokens at once, even though supervising early depths can encourage their hidden states to carry information about future tokens~\citep{pal2023future, elhoushi2024layerskip}, as we demonstrate in \Cref{sec:dense}.

\paragraph{Speculative decoding.}
Speculative decoding \citep{leviathan2023fast, chen2023accelerating} accelerates a target model $p$ using a faster draft model $q$ that proposes upcoming tokens. Given a prefix ${\bf x}^{\le t}$, the drafter proposes $k$ draft tokens ${\bf d}=({\bf d}^1,\dots,{\bf d}^k)\in V^k$, and the target model evaluates all of them in a single decoding step to obtain its own predictions. Under greedy decoding, this is done as:
\begin{equation}
\label{eq:verify}
{\bf y}^j=\argmax({\bf p}^j),\qquad {\bf p}^j\coloneqq p(\cdot\mid{\bf x}^{\le t},{\bf d}^{<j}),\qquad j=1,\dots,k+1,
\end{equation}
and the drafts are accepted up to the first mismatch ${\bf d}^j\neq{\bf y}^j$, where the target model substitutes its own prediction ${\bf y}^j$. Each step thus produces at least one and at most $k+1$ tokens while keeping the output identical to greedy decoding with $p$. Under sampling, each draft ${\bf d}^j$ is drawn from a draft distribution ${\bf q}^j$ and accepted with probability $\min\{1,\,{\bf p}^j({\bf d}^j)/{\bf q}^j({\bf d}^j)\}$. Upon the first rejection, a replacement token is sampled from the normalization of $\max\{0,\,{\bf p}^j-{\bf q}^j\}$, which leaves the output distribution of $p$ unchanged \citep{leviathan2023fast}. Because verifying $k+1$ tokens in parallel at small batch sizes takes nearly the same time as a single-token decoding step, the speedup grows with the number of accepted tokens $\tauacc$ generated per step and decreases with the relative cost of drafting.

\section{Floodgates}
\label{sec:method}

In this section, we present \ours, a simple self-speculative decoding framework for looped language models. \Cref{alg:train,alg:decode} summarize training and decoding.

\paragraph{Drafting.}
Let ${\bf x}^t$ be the latest generated token and ${\bf m}\in\mathbb{R}^d$ be a learnable vector. To draft $k$ tokens in a single loop iteration, we append $k$ \emph{mask tokens} after ${\bf x}^t$ by repeating ${\bf m}$, forming $k+1$ positions that we call slots, and run the shared block $f_\theta$ once over all slots in parallel while attending to the \emph{deep prefix cache} ${\bf c}^{<t}_R$. We then read out the resulting slot states with the head $g$ to obtain the draft distributions ${\bf q}^j$:
\begin{equation}
\label{eq:draft}
(\hat{\bf z}^t,\hat{\bf z}^{t+1},\dots,\hat{\bf z}^{t+k}) = f_{\theta}\bigl(({\bf z}^t_0,{\bf m},\dots,{\bf m});\,{\bf c}^{<t}_R\bigr), \qquad {\bf q}^j = g(\hat{\bf z}^{t+j}), \quad j=1,\dots,k,
\end{equation}
where $\hat{\bf z}^{t+j}\in\mathbb{R}^d$ is the hidden state of slot $j$, and ${\bf d}^j=\argmax({\bf q}^j)$ under greedy decoding, for instance. While mask tokens have been used for parallel drafting with dense (non-looped) models~\citep{monea2023pass,bhendawade2024streaming,samragh2025your,sahoo2026uno}, a dense model must pass the mask tokens through all of its layers. In our approach, the slots pass through the shared block only once, using all parameters of the model at the cost of a single loop, and they attend to the deep prefix cache ${\bf c}^{<t}_R$ instead of the first-loop cache ${\bf c}^{<t}_1$. Here, ${\bf c}^{<t}_R$ provides the model's full-depth representation of all earlier tokens ${\bf x}^{<t}$, which is already in the cache from verifying tokens from previous steps. To teach a single loop to predict these tokens without changing the verifier, we use lightweight adapters that are active only while drafting.

\begin{figure}[t!]
\begin{minipage}[t]{0.47\textwidth}
\algrenewcommand{\algorithmicindent}{0.6em}
\begin{algorithm}[H]
\caption{Training}
\label{alg:train}
\begin{algorithmic}[1]
\small
\Require Looped block $f_\theta$, head $g$, draft length $k$, weights $w_{1:k}$, learning rate $\eta$, initial adapters $\phi$ and mask ${\bf m}$
\vspace*{0.5\baselineskip}
\Repeat
    \State ${\bf x}\sim\mathcal{D}$ \Comment{self-generated}
    \State $t\sim\mathrm{Unif}\{1,\dots,|{\bf x}|\}$
    \State run \cref{eq:loop} on ${\bf x}^{<t}$ to obtain ${\bf c}^{<t}_R$ \Comment{prefix cache}
    \State $\text{slots}\gets({\bf z}^t_0,{\bf m},\dots,{\bf m})$
    \State $(\hat{\bf z}^t,\dots,\hat{\bf z}^{t+k})\gets f_\theta(\text{slots};\,{\bf c}^{<t}_R)$ with $\phi$ \Comment{draft}
    \State ${\bf q}^j\gets g(\hat{\bf z}^{t+j})$ for $j\le k$
    \State $\mathcal{L}\gets\sum_{j=1}^{k} w_j\,\mathsf{CE}({\bf q}^j,{\bf x}^{t+j})$ \Comment{\cref{eq:loss}}
    \State $(\phi,{\bf m})\gets(\phi,{\bf m})-\eta\,\nabla_{\phi,{\bf m}}\mathcal{L}$ \Comment{$\theta$ frozen}
\Until{converged}
\State \algreturn $(\phi,{\bf m})$
\end{algorithmic}
\vspace*{0.5\baselineskip}
\end{algorithm}
\end{minipage}%
\hfill
\begin{minipage}[t]{0.51\textwidth}
\algrenewcommand{\algorithmicindent}{0.6em}
\begin{algorithm}[H]
\caption{Decoding step}
\label{alg:decode}
\begin{algorithmic}[1]
\small
\Require Prefix ${\bf x}^{\le t}$, deep prefix cache ${\bf c}^{<t}_R$, model $(f_\theta,g)$, drafter $(\phi,{\bf m})$, draft length $k$, elastic draft budget $E$, window $W$
\State $\text{slots}\gets({\bf z}^t_0,{\bf m},\dots,{\bf m})$
\State $(\hat{\bf z}^t,\dots,\hat{\bf z}^{t+k})\gets f_\theta(\text{slots};\,{\bf c}^{<t}_R)$ with $\phi$ \Comment{draft}
\State ${\bf q}^j\gets g(\hat{\bf z}^{t+j})$; ${\bf d}^j\gets\argmax({\bf q}^j)$ for $j\le k$
\State ${\bf d}^j\gets{\bf d}^k$ and $\hat{\bf z}^{t+j}\gets{\bf z}^{t+j}_0$ for $k<j\le W$ \Comment{extend}
\For{$e=1,\dots,E$}
\State $(\hat{\bf z}^t,\dots,\hat{\bf z}^{t+W})\gets f_\theta\bigl((\hat{\bf z}^{t+j},{\bf E}_{{\bf d}^j})_{j=0}^{W};\,{\bf c}^{<t}_R\bigr)$ \Comment{\cref{eq:refine}}
\State ${\bf d}^j\gets\argmax(g(\hat{\bf z}^{t+j}))$ for $j\le W$
\EndFor
\State ${\bf p}^j\gets p(\cdot\mid{\bf x}^{\le t},{\bf d}^{<j})$ for $j\le W+1$ \Comment{verify}
\State ${\bf y}^j\gets\argmax({\bf p}^j)$ for $j\le W+1$
\State $a\gets\max\{\,j : {\bf d}^1={\bf y}^1,\dots,{\bf d}^j={\bf y}^j\,\}$ \Comment{accept}
\State \algreturn $({\bf x}^{\le t},{\bf d}^1,\dots,{\bf d}^a,\,{\bf y}^{a+1})$
\end{algorithmic}
\end{algorithm}
\end{minipage}
\end{figure}

\paragraph{Training.}
\label{sec:training}
Our insight is that all-depth training (\ref{eq:loop_loss}) already encourages the shared block $f_\theta$ to build features that are predictive of several future tokens in its early loops (see \Cref{sec:dense}). During pretraining, however, the first loop is trained to output only the immediate next token and attends only to the first-loop cache ${\bf c}^{<t}_1$. Therefore, our approach further learns to read the deep prefix cache ${\bf c}^{<t}_R$ and decode those future tokens into the mask slots. To this end, we train the slots with a cross-entropy objective applied $j$ tokens ahead:
\begin{equation}
\label{eq:loss}
\mathcal{L}(\phi,{\bf m}) = \mathbb{E}_{{\bf x},t}\Bigl[\,\sum_{j=1}^{k} w_j\,\mathsf{CE}\bigl(g(\hat{\bf z}^{t+j}),\,{\bf x}^{t+j}\bigr)\Bigr],
\end{equation}
where ${\bf x}\sim\mathcal{D}$ is a sequence generated by the target model itself and $t$ is a sampled position in its continuation after which the slots are placed. We train on the model's own generations~\citep{cai2024medusa} because verification (\ref{eq:verify}) checks drafts against the model predictions.  Because a draft at slot $j$ can be accepted only if all earlier drafts $<j$ are also accepted, we set exponentially decaying weights $w_j \propto e^{-(j-1)/4}$ normalized to sum to one to incentivize the drafter to prioritize accuracy on earlier positions. We parameterize the drafter with low-rank adapters $\phi$ \citep{hu2022lora} that are active only at the slots \citep{samragh2025your}, keeping the base model weights $\theta$ and output head $g$ frozen.

\paragraph{Elastic drafting.}
\label{sec:refine}
When a looped model runs many loops, either by design or through test-time scaling, a single loop may be too shallow to draft a long continuation that the full model will accept. We find that the pretrained loop $f_\theta$ can be applied out of the box to iteratively refine the initial draft over a window of $W\ge k$ positions. Specifically, each position $j\in\{0,\dots,W\}$ in the window maintains a hidden feature $\hat{\bf z}^{t+j}\in\mathbb{R}^d$ and a draft token ${\bf d}^j$, with ${\bf d}^0={\bf x}^t$. Before the first pass, the first $k$ positions ($1\le j\le k$) take their features $\hat{\bf z}^{t+j}$ and drafted tokens ${\bf d}^j$ directly from the single-loop draft of (\ref{eq:draft}), while the remaining $W-k$ extended positions ($k<j\le W$) repeat the last drafted token ${\bf d}^j\gets {\bf d}^k$ and initialize $\hat{\bf z}^{t+j}$ to its initial state ${\bf z}^{t+j}_0$. We then run $E\ge 0$ elastic passes that jointly update the features and draft tokens across the window:
\begin{equation}
\label{eq:refine}
(\hat{\bf z}^t,\dots,\hat{\bf z}^{t+W}) \gets f_{\theta}\bigl(\bigl(\hat{\bf z}^{t+j},{\bf E}_{{\bf d}^j}\bigr)_{j=0}^{W};\,{\bf c}^{<t}_R\bigr), \qquad {\bf d}^j \gets \argmax\bigl(g(\hat{\bf z}^{t+j})\bigr),
\end{equation}
where ${\bf E}_{{\bf d}^j}\in\mathbb{R}^d$ is the token embedding of ${\bf d}^j$, combined with the feature $\hat{\bf z}^{t+j}$ through the model's input projection so that each pass revises both the features and the draft tokens using the latest tokens in the window. These passes do not use the adapters $\phi$ and therefore require no extra training. While this update resembles Jacobi decoding \citep{santilli2023accelerating, fu2024break} and DAS-Wave \citep{li2026shallow}, which refine uninitialized or repeated tokens from scratch, elastic drafting starts from the trained multi-token draft of (\ref{eq:draft}) and therefore requires far fewer passes.

Iterating (\ref{eq:refine}) creates a direct tradeoff between drafting cost and acceptance, since running more passes $E$ improves the draft at the cost of one additional loop per pass. Because each pass updates all $W$ draft tokens in parallel, running $E$ passes costs only $E/W$ effective loops per position, which is much cheaper than the $R$ loops of tokenwise generation and yields higher speedup when $E$ and $W$ are chosen adaptively to the verifier depth $R$. On shallow models with, e.g., $R=4$, where a single loop already costs a quarter of a decoding step, single-pass drafting with $E=0$ is fastest, whereas on deep models with, e.g., $R=32$, where each loop is a small fraction of a step, additional passes $E>0$ pay off. We treat $E$ as a hyperparameter and select the value that maximizes speedup on held-out problems for each model (\Cref{sec:app_passes}), following standard practice for choosing the draft block size in speculative decoding \citep{leviathan2023fast, chen2024sequoia}.

\paragraph{Concurrent serving.}
\label{sec:method_batch}
Because an untrained shallow loop only predicts the immediate next token, prior self-speculative methods for looped models~\citep{cho2026loopspec,ha2026wavefront} rely on \emph{depth pipelining}, starting the next token at the first loop before earlier tokens have finished their $R$ loops. While pipelining parallelizes drafting and verification, it batches together tokens at different loop depths, say $B\subseteq[R]$ (e.g., one every four loops), and therefore must store and read the prefix cache ${\bf c}^{<t}_r$ of all such loops $r\in B$.
When serving many concurrent requests, maintaining and reading $B$ separate caches per request saturates runtime memory bandwidth, causing diminishing return in throughput.
By drafting a block of future tokens in parallel from a single loop, \ours avoids pipelining altogether. All drafted slots of a sequence attend only to the deep prefix cache ${\bf c}^{<t}_R$, reading the keys and values of past tokens once per step for the entire draft block.

At batch size one, decoding is memory-bound, so verifying all $k$ draft tokens in parallel takes nearly the same time as verifying a single token. As the batch size $B$ grows, verification becomes compute-bound and its runtime increases with the total number of verified drafts, so checking all $kB$ candidates across the batch wastes compute on low-confidence tokens at later positions~\citep{sahoo2026uno}. To keep throughput high across batch sizes, we use single-pass drafting ($E=0$) and verify only the most promising drafts in the batch. Specifically, let $\alpha_{b,j}$ be the probability ${\bf q}^j({\bf d}^j)$ assigned by the drafter to slot $j$ of sequence $b$. Because slot $j$ is accepted only if all earlier slots $i\le j$ in sequence $b$ are also accepted, we estimate its acceptance probability as $\pi_{b,j}=\prod_{i=1}^{j}\alpha_{b,i}$. Given a total budget of $M$ drafts across the batch, we select the subset $\mathcal{S}_M$ of $M$ slots with the largest $\pi_{b,j}$; verifying $\mathcal{S}_M$ takes time $T(M)$ and yields $B + \sum_{(b,j)\in\mathcal{S}_M} \pi_{b,j}$ tokens in expectation ($B$ guaranteed tokens from verification plus the expected accepted drafts). At each step, we therefore choose the budget $M^\star$ that maximizes expected throughput:
\begin{equation}
\label{eq:batch_select}
M^\star = \argmax_{M\in\{B,\,2B,\,\dots,\,kB\}}\;\frac{B + \sum_{(b,j)\in\mathcal{S}_M} \pi_{b,j}}{T(M)},
\end{equation}
where the verification time $T(M)$ is profiled once at initialization for multiples $M$ of $B$. Since $\pi_{b,j}$ decreases with $j$, the selected slots form a prefix of the draft block of each sequence. The remaining drafts are discarded without running further loops, and each sequence drafts a new block after its last verified token at the next step.

\section{Experiments}
\label{sec:experiments}

Through experiments, we aim to answer the following core questions:
\begin{enumerate}[label={\bf Q\arabic*}.,leftmargin=*,itemsep=1pt,topsep=2pt]
\item Can \ours accelerate looped decoding across models and decoding regimes (\Cref{sec:main_results})?
\item How do all-depth supervision and deep prefix conditioning contribute to multi-token drafting (\Cref{sec:dense})?
\item Does elastic drafting adapt its depth to maximize performance (\Cref{sec:controls})?
\item Do the decoding gains persist under high-throughput batched serving (\Cref{sec:batch})?
\end{enumerate}

The dataset, hyperparameters, and hardware configurations for each model can be found in \Cref{sec:app_training}.

\subsection{Accelerating looped decoding across models and regimes (Q1)}
\label{sec:main_results}
\begin{table}[!t]
\centering
\caption{\textbf{\ours is the fastest lossless method on every model and task under greedy decoding, and in 16 of 24 settings under sampling.} We report wall-clock speedup ($\times$) over plain decoding on an A100 at batch size one, with accepted tokens per step $\tauacc$ in parentheses. Sampling uses temperature $1.0$ and top-$p$ $0.7$. We bold the fastest lossless method and underline the second fastest. LoopSpec$^\ast$ uses one proposal~\citep{cho2026loopspec}, WaveFront~\citep{ha2026wavefront} and DAS/DAS-Wave~\citep{li2026shallow} are training-free looped baselines, JD denotes Jacobi decoding~\citep{santilli2023accelerating}, and PLD denotes prompt lookup decoding~\citep{saxena2023pld}.}
\label{tab:main}
\fitwidth{%
\setlength{\tabcolsep}{2.2pt}
\begin{tabular}{llcccccccccc}
\toprule
 & & \multicolumn{2}{c}{GSM8K} & \multicolumn{2}{c}{MATH-500} & \multicolumn{2}{c}{HumanEval+} & \multicolumn{2}{c}{MBPP+} & \multicolumn{2}{c}{BBH} \\
\cmidrule(lr){3-4}\cmidrule(lr){5-6}\cmidrule(lr){7-8}\cmidrule(lr){9-10}\cmidrule(lr){11-12}
Model & Method & Greedy & Sampling & Greedy & Sampling & Greedy & Sampling & Greedy & Sampling & Greedy & Sampling \\
\midrule
\multirow{8}{*}{\makecell[l]{Ouro\\1.4B}} & LoopSpec$^\ast$ & 2.51 (3.1) & 2.10 (2.8) & 2.60 (3.2) & 2.34 (3.0) & 3.01 (3.7) & 2.67 (3.5) & 2.62 (3.3) & 2.23 (3.0) & 2.67 (3.3) & 2.25 (2.9) \\
 & LoopSpec & \underline{2.77} (3.5) & \underline{2.40} (3.3) & \underline{2.81} (3.5) & \underline{2.53} (3.4) & 3.10 (3.9) & 2.77 (3.8) & \underline{2.83} (3.6) & \underline{2.51} (3.5) & 2.86 (3.6) & 2.46 (3.4) \\
 & WaveFront & 2.34 (3.1) & 1.85 (2.4) & 2.47 (3.2) & 2.10 (2.7) & 2.77 (3.7) & 2.31 (3.0) & 2.38 (3.2) & 2.00 (2.6) & 2.52 (3.4) & 1.96 (2.5) \\
 & DAS & 1.91 (5.8) & 1.74 (5.3) & 2.07 (6.4) & 1.97 (6.0) & 2.59 (7.9) & 2.40 (7.3) & 2.06 (6.3) & 1.83 (5.6) & 2.16 (6.6) & 2.07 (6.3) \\
 & DAS-Wave & 0.44 (1.8) & 0.54 (1.9) & 0.44 (1.8) & 0.53 (1.9) & 0.43 (1.8) & 0.52 (1.8) & 0.44 (1.8) & 0.50 (1.8) & 0.42 (1.7) & 0.50 (1.8) \\
 & JD & 0.95 (3.3) & 0.83 (4.6) & 0.94 (5.3) & 0.87 (4.9) & 0.90 (5.1) & 0.87 (4.9) & 0.91 (3.2) & 0.78 (4.4) & 0.93 (5.3) & 0.90 (5.0) \\
 & PLD & 2.08 (2.0) & 1.97 (1.9) & 1.96 (2.0) & 1.86 (1.9) & \underline{3.27} (3.3) & \underline{2.91} (2.9) & 1.37 (1.4) & 1.33 (1.3) & \underline{3.38} (3.4) & \underline{3.23} (3.2) \\
 & \cellcolor{LoopedFlowLight}\ours & \cellcolor{LoopedFlowLight}\textbf{5.04} (7.5) & \cellcolor{LoopedFlowLight}\textbf{4.16} (6.3) & \cellcolor{LoopedFlowLight}\textbf{4.69} (7.0) & \cellcolor{LoopedFlowLight}\textbf{4.41} (6.7) & \cellcolor{LoopedFlowLight}\textbf{4.83} (7.2) & \cellcolor{LoopedFlowLight}\textbf{4.49} (6.8) & \cellcolor{LoopedFlowLight}\textbf{3.01} (4.6) & \cellcolor{LoopedFlowLight}\textbf{2.75} (4.3) & \cellcolor{LoopedFlowLight}\textbf{4.40} (6.6) & \cellcolor{LoopedFlowLight}\textbf{4.11} (6.2) \\
\midrule
\multirow{8}{*}{\makecell[l]{Ouro\\2.6B}} & LoopSpec$^\ast$ & 2.86 (3.4) & 2.68 (3.3) & 2.90 (3.4) & 2.79 (3.4) & 3.28 (3.8) & 3.03 (3.6) & 2.94 (3.5) & 2.79 (3.3) & 2.95 (3.5) & 2.65 (3.2) \\
 & LoopSpec & \underline{3.06} (3.6) & \underline{2.89} (3.6) & \underline{3.10} (3.7) & \underline{2.96} (3.6) & \underline{3.33} (3.9) & \underline{3.14} (3.8) & \underline{3.09} (3.7) & \underline{2.95} (3.7) & 3.10 (3.7) & 2.86 (3.6) \\
 & WaveFront & 2.53 (3.3) & 2.01 (2.6) & 2.50 (3.3) & 2.20 (2.9) & 2.83 (3.8) & 2.32 (3.0) & 2.51 (3.3) & 2.18 (2.9) & 2.61 (3.5) & 2.13 (2.8) \\
 & DAS & 2.24 (6.7) & 2.04 (6.1) & 2.35 (7.1) & 2.22 (6.7) & 2.71 (8.2) & 2.47 (7.4) & 2.19 (6.6) & 1.96 (5.9) & 2.36 (7.1) & 2.28 (6.8) \\
 & DAS-Wave & 0.55 (2.2) & 0.66 (2.3) & 0.49 (2.0) & 0.56 (2.0) & 0.51 (2.0) & 0.59 (2.1) & 0.50 (2.1) & 0.58 (2.1) & 0.49 (2.0) & 0.59 (2.1) \\
 & JD & 0.95 (3.2) & 0.81 (4.5) & 0.94 (5.3) & 0.87 (4.9) & 0.90 (5.1) & 0.85 (4.8) & 0.86 (3.0) & 0.75 (4.3) & 0.91 (5.2) & 0.88 (5.0) \\
 & PLD & 2.04 (2.0) & 1.94 (1.9) & 2.79 (2.7) & 2.75 (2.7) & 3.17 (3.1) & 2.55 (2.5) & 1.31 (1.3) & 1.34 (1.3) & \underline{3.40} (3.4) & \underline{3.28} (3.3) \\
 & \cellcolor{LoopedFlowLight}\ours & \cellcolor{LoopedFlowLight}\textbf{5.56} (8.2) & \cellcolor{LoopedFlowLight}\textbf{4.50} (6.7) & \cellcolor{LoopedFlowLight}\textbf{6.04} (9.0) & \cellcolor{LoopedFlowLight}\textbf{4.90} (7.3) & \cellcolor{LoopedFlowLight}\textbf{5.57} (8.3) & \cellcolor{LoopedFlowLight}\textbf{4.82} (7.2) & \cellcolor{LoopedFlowLight}\textbf{3.39} (5.2) & \cellcolor{LoopedFlowLight}\textbf{3.23} (4.9) & \cellcolor{LoopedFlowLight}\textbf{5.07} (7.5) & \cellcolor{LoopedFlowLight}\textbf{4.77} (7.1) \\
\midrule
\multirow{7}{*}{\makecell[l]{Recurrent\\Llama 3.2}} & LoopSpec$^\ast$ & 3.27 (7.7) & 2.98 (7.5) & 5.03 (11.3) & 3.86 (9.3) & 5.14 (11.7) & 4.10 (10.0) & 3.91 (9.1) & 3.79 (9.2) & 4.27 (10.1) & 3.86 (9.6) \\
 & LoopSpec & \underline{4.61} (12.2) & \textbf{3.80} (12.3) & \underline{6.11} (14.5) & \textbf{4.73} (13.5) & \underline{6.03} (14.5) & \textbf{4.81} (13.9) & \underline{5.41} (13.5) & \textbf{4.63} (13.4) & 5.65 (13.8) & \underline{4.77} (13.7) \\
 & DAS & 3.82 (7.5) & 2.77 (5.4) & 4.30 (8.4) & 3.45 (6.7) & 4.32 (8.4) & 3.58 (7.0) & 4.01 (7.8) & 3.38 (6.6) & 4.22 (8.2) & 3.83 (7.5) \\
 & DAS-Wave & 4.56 (17.9) & 2.78 (6.1) & 5.98 (23.6) & 3.71 (14.7) & 5.92 (23.1) & 3.95 (15.4) & 5.19 (20.3) & 3.52 (13.7) & \underline{5.84} (23.1) & 4.63 (18.2) \\
 & JD & 0.76 (4.9) & 0.68 (2.6) & 0.80 (3.2) & 0.71 (2.8) & 0.80 (3.1) & 0.74 (2.9) & 0.78 (3.1) & 0.74 (2.9) & 0.81 (3.2) & 0.78 (3.1) \\
 & PLD & 1.83 (1.9) & 1.55 (1.6) & 3.74 (4.0) & 2.66 (2.8) & 3.54 (3.7) & 3.03 (3.2) & 2.23 (2.3) & 2.31 (2.4) & 3.48 (3.7) & 3.24 (3.5) \\
 & \cellcolor{LoopedFlowLight}\ours & \cellcolor{LoopedFlowLight}\textbf{5.04} (20.4) & \cellcolor{LoopedFlowLight}\underline{3.28} (7.6) & \cellcolor{LoopedFlowLight}\textbf{6.97} (28.5) & \cellcolor{LoopedFlowLight}\underline{4.66} (11.0) & \cellcolor{LoopedFlowLight}\textbf{6.91} (28.0) & \cellcolor{LoopedFlowLight}\underline{4.69} (10.9) & \cellcolor{LoopedFlowLight}\textbf{5.84} (23.5) & \cellcolor{LoopedFlowLight}\underline{4.34} (10.1) & \cellcolor{LoopedFlowLight}\textbf{6.62} (27.1) & \cellcolor{LoopedFlowLight}\textbf{5.13} (12.1) \\
\midrule
\multirow{8}{*}{\makecell[l]{Huginn\\3.5B}} & LoopSpec$^\ast$ & 1.35 (2.4) & 1.14 (2.1) & 2.54 (5.3) & 1.44 (2.9) & 2.20 (3.9) & 1.76 (3.2) & 1.87 (3.2) & 1.69 (3.0) & 1.48 (3.0) & 1.20 (2.5) \\
 & LoopSpec & 2.20 (5.4) & 1.64 (4.7) & 3.07 (9.0) & 2.16 (7.4) & 3.87 (9.0) & 2.89 (7.8) & 3.21 (7.3) & 2.62 (6.7) & 2.02 (7.2) & 1.43 (6.3) \\
 & WaveFront & 2.29 (3.1) & 1.62 (2.5) & 2.99 (5.3) & 1.76 (3.0) & 3.53 (5.8) & 2.27 (3.6) & 2.85 (4.6) & 2.03 (3.2) & 2.55 (4.4) & 1.96 (3.3) \\
 & DAS & 2.74 (6.2) & 2.20 (5.0) & 3.47 (7.9) & 2.88 (6.6) & 3.59 (8.2) & 3.15 (7.2) & 3.05 (6.9) & 2.57 (5.9) & 3.38 (7.7) & 3.06 (7.0) \\
 & DAS-Wave & \underline{4.23} (17.6) & \underline{2.57} (5.7) & \underline{5.38} (22.8) & \underline{3.57} (15.0) & \underline{5.45} (22.7) & \underline{4.15} (17.3) & \underline{4.57} (19.0) & \underline{3.02} (6.7) & \underline{5.20} (22.0) & \underline{4.22} (17.8) \\
 & JD & 0.83 (3.1) & 0.74 (2.8) & 0.82 (3.1) & 0.77 (3.0) & 0.80 (3.1) & 0.80 (3.0) & 0.81 (3.1) & 0.74 (2.8) & 0.81 (3.1) & 0.78 (3.0) \\
 & PLD & 2.11 (2.2) & 1.88 (1.9) & 4.16 (4.3) & 3.51 (3.7) & 4.77 (4.9) & 3.51 (3.6) & 2.26 (2.3) & 2.08 (2.1) & 3.43 (3.6) & 3.28 (3.4) \\
 & \cellcolor{LoopedFlowLight}\ours & \cellcolor{LoopedFlowLight}\textbf{4.48} (19.0) & \cellcolor{LoopedFlowLight}\textbf{3.16} (7.3) & \cellcolor{LoopedFlowLight}\textbf{6.21} (26.6) & \cellcolor{LoopedFlowLight}\textbf{4.60} (10.8) & \cellcolor{LoopedFlowLight}\textbf{6.18} (26.2) & \cellcolor{LoopedFlowLight}\textbf{4.70} (10.9) & \cellcolor{LoopedFlowLight}\textbf{5.06} (21.4) & \cellcolor{LoopedFlowLight}\textbf{3.52} (8.1) & \cellcolor{LoopedFlowLight}\textbf{5.70} (24.6) & \cellcolor{LoopedFlowLight}\textbf{4.48} (19.3) \\
\midrule
\multirow{7}{*}{\makecell[l]{Recurrent\\TinyLlama}} & LoopSpec$^\ast$ & 3.92 (7.7) & 3.47 (7.2) & 5.39 (10.2) & 4.70 (9.2) & 6.25 (11.8) & 4.71 (9.5) & 4.97 (9.7) & 4.38 (8.9)  & \multicolumn{2}{c}{\multirow{7}{*}{\makecell{prompts exceed\\max context length}}}\\
 & LoopSpec & 5.79 (12.6) & \textbf{4.95} (12.4) & \underline{6.92} (13.9) & \textbf{5.90} (13.3) & \underline{7.35} (14.7) & \textbf{6.13} (13.9) & \underline{6.48} (13.7) & \textbf{5.89} (13.6)  & \multicolumn{2}{c}{}\\
 & DAS & 4.20 (7.6) & 3.06 (5.5) & 4.53 (8.2) & 3.57 (6.4) & 4.66 (8.4) & 3.93 (7.1) & 4.46 (8.0) & 3.75 (6.8)  & \multicolumn{2}{c}{}\\
 & DAS-Wave & \underline{6.07} (19.2) & 3.40 (6.5) & 6.86 (22.1) & 3.82 (12.2) & 7.10 (22.6) & 4.81 (15.2) & 6.43 (20.4) & 4.13 (13.1)  & \multicolumn{2}{c}{}\\
 & JD & 0.87 (3.2) & 0.77 (2.9) & 0.86 (3.3) & 0.77 (2.9) & 0.84 (3.1) & 0.79 (2.9) & 0.85 (3.2) & 0.80 (3.0)  & \multicolumn{2}{c}{}\\
 & PLD & 1.98 (2.1) & 1.77 (1.9) & 2.76 (2.9) & 2.34 (2.5) & 3.86 (4.1) & 2.85 (3.0) & 2.14 (2.2) & 2.29 (2.4)  & \multicolumn{2}{c}{}\\
 & \cellcolor{LoopedFlowLight}\ours & \cellcolor{LoopedFlowLight}\textbf{6.73} (22.2) & \cellcolor{LoopedFlowLight}\underline{4.01} (8.1) & \cellcolor{LoopedFlowLight}\textbf{7.79} (26.0) & \cellcolor{LoopedFlowLight}\underline{4.83} (9.9) & \cellcolor{LoopedFlowLight}\textbf{8.20} (27.0) & \cellcolor{LoopedFlowLight}\underline{5.61} (11.4) & \cellcolor{LoopedFlowLight}\textbf{7.17} (23.5) & \cellcolor{LoopedFlowLight}\underline{5.25} (10.6)  & \multicolumn{2}{c}{\cellcolor{LoopedFlowLight}}\\
\bottomrule
\end{tabular}}
\end{table}

\begin{table}[!tp]
\centering
\caption{\textbf{On the reasoning models, \ours is the fastest method on both tasks under greedy decoding and sampling.} We report wall-clock speedup ($\times$) over plain decoding on an A100 at batch size one, with accepted tokens per step $\tauacc$ in parentheses. Sampling uses temperature $0.6$ and top-$p$ $0.95$. We bold the fastest lossless method and underline the second fastest.}
\label{tab:thinking}
\footnotesize
\begin{tabular}{llcccc}
\toprule
 & & \multicolumn{2}{c}{GSM8K} & \multicolumn{2}{c}{MATH-500} \\
\cmidrule(lr){3-4}\cmidrule(lr){5-6}
Model & Method & Greedy & Sampling & Greedy & Sampling \\
\midrule
\multirow{7}{*}{Ouro 1.4B Thinking} & LoopSpec & \underline{2.72} (3.3) & \underline{2.40} (3.3) & \underline{2.74} (3.5) & \underline{2.47} (3.5) \\
 & WaveFront & 2.24 (2.9) & 2.13 (2.8) & 2.31 (3.2) & 2.18 (3.0) \\
 & DAS & 1.76 (5.5) & 1.65 (5.1) & 1.97 (6.0) & 1.80 (5.5) \\
 & DAS-Wave & 0.47 (1.8) & 0.47 (1.8) & 0.48 (1.9) & 0.46 (1.8) \\
 & JD & 0.86 (5.3) & 0.77 (4.7) & 0.85 (5.3) & 0.76 (4.8) \\
 & PLD & 1.65 (1.7) & 1.59 (1.6) & 2.29 (2.3) & 1.81 (1.9) \\
 & \cellcolor{LoopedFlowLight}\ours & \cellcolor{LoopedFlowLight}\textbf{3.70} (5.6) & \cellcolor{LoopedFlowLight}\textbf{3.46} (5.2) & \cellcolor{LoopedFlowLight}\textbf{3.68} (5.5) & \cellcolor{LoopedFlowLight}\textbf{3.23} (4.9) \\
\midrule
\multirow{7}{*}{Ouro 2.6B Thinking} & LoopSpec & \underline{3.09} (3.5) & \underline{2.87} (3.6) & \underline{2.98} (3.6) & \underline{2.85} (3.6) \\
 & WaveFront & 2.42 (3.1) & 2.39 (3.1) & 2.42 (3.3) & 2.36 (3.2) \\
 & DAS & 2.05 (6.2) & 1.89 (5.8) & 2.23 (6.8) & 2.02 (6.1) \\
 & DAS-Wave & 0.57 (2.2) & 0.56 (2.1) & 0.51 (2.0) & 0.50 (2.0) \\
 & JD & 0.86 (5.3) & 0.76 (4.7) & 0.87 (5.5) & 0.77 (4.8) \\
 & PLD & 1.62 (1.7) & 1.60 (1.6) & 2.03 (2.1) & 1.84 (1.9) \\
 & \cellcolor{LoopedFlowLight}\ours & \cellcolor{LoopedFlowLight}\textbf{3.90} (5.9) & \cellcolor{LoopedFlowLight}\textbf{3.71} (5.5) & \cellcolor{LoopedFlowLight}\textbf{3.72} (5.5) & \cellcolor{LoopedFlowLight}\textbf{3.41} (5.1) \\
\bottomrule
\end{tabular}
\end{table}

\paragraph{Setup.}
\label{sec:setup}
We evaluate \ours on seven looped models from two families. Ouro~1.4B and Ouro~2.6B~\citep{zhu2025scaling}, together with their reasoning variants Ouro~1.4B~Thinking and Ouro~2.6B~Thinking, loop their entire stack of layers $R=4$ times. Recurrent~Llama~3.2, Recurrent~TinyLlama~\citep{mcleish2025teaching}, and Huginn 3.5B~\citep{geiping2025scaling} run their initial layers (``prelude'') once, loop their middle layers (``core'') $R=32$ times by default, and run their final layers (``coda'') once. Because these three models were trained with randomized loop counts, they can be run at any depth at inference time. We evaluate on five standard benchmarks spanning mathematical reasoning on GSM8K~\citep{cobbe2021training} and MATH-500~\citep{hendrycks2021measuring,lightman2024lets}, code generation on HumanEval+ and MBPP+~\citep{chen2021evaluating,austin2021program,liu2023evalplus}, and general reasoning on BBH~\citep{suzgun2023challenging}, under both greedy decoding and sampling following the protocol of LoopSpec~\citep{cho2026loopspec}. Using 100 held-out problems, we set the elastic drafting budget $E$ as a single loop iteration, $E=0$, on 4-loop Ouro models, and $E=24$ on the remaining 32-loop models. We leave a comparison of $E=0$ and $24$ across every model and task in \Cref{sec:app_passes} for reference.

We compare \ours with existing training-free self-speculative decoding methods for looped models, namely LoopSpec~\citep{cho2026loopspec}, WaveFront~\citep{ha2026wavefront}, DAS, and DAS-Wave~\citep{li2026shallow}, as well as Jacobi decoding (JD)~\citep{santilli2023accelerating,fu2024break}, and prompt lookup decoding (PLD)~\citep{saxena2023pld}.
We report wall-clock speedup over plain decoding and the average number of accepted tokens per step $\tauacc$ on one NVIDIA A100-80GB GPU at batch size one. Full experimental details are provided in \Cref{sec:app_details}.

\paragraph{Results.}
\Cref{tab:main,tab:thinking} show the results. Under greedy decoding, \ours is the fastest lossless method on every model and task, with the largest margin on the shallow, 4-loop Ouro models (Ouro~1.4B and Ouro~2.6B). There, \ours drafts with a single loop iteration without elastic passes (\Cref{sec:method}) to accept $4.6$--$9.0$ tokens per step, decoding $3.0$--$6.0\times$ faster than plain decoding while no prior method exceeds $3.4\times$. On the deeper, 32-loop models (Recurrent~Llama~3.2, Huginn 3.5B, and Recurrent~TinyLlama), \ours uses elastic drafting with $24$ elastic passes (\Cref{sec:refine}) to accept between $19$ and $29$ tokens per step, which reaches $4.5$--$8.2\times$ speedup and outperforms the strongest training-free baselines, LoopSpec and DAS-Wave, by $6$--$15\%$ on every task.

Under sampling, stochastic verification at each draft position reduces the number of accepted tokens per step for all block drafters. Nevertheless, in \Cref{tab:main}, \ours remains the fastest lossless method across all five benchmarks on Ouro~1.4B ($2.8$--$4.5\times$), Ouro~2.6B ($3.2$--$4.9\times$), and Huginn 3.5B ($3.2$--$4.7\times$), as well as on BBH for Recurrent~Llama~3.2 ($5.1\times$), while LoopSpec is faster under sampling on Recurrent~TinyLlama and four tasks of Recurrent~Llama~3.2. On the long-horizon reasoning models Ouro~1.4B~Thinking and Ouro~2.6B~Thinking (\Cref{tab:thinking}), \ours achieves the highest speedup on every task under both greedy decoding ($3.7$--$3.9\times$ vs.\ $2.7$--$3.1\times$ for LoopSpec) and sampling ($3.2$--$3.7\times$ vs.\ $2.4$--$2.9\times$). In \Cref{sec:app_more}, we provide confidence intervals for all speedups, verify that decoding is lossless through task accuracy measurements, and further compare \ours against lossy acceleration methods.

\subsection{Contribution of all-depth supervision and deep prefix conditioning (Q2)}
 \label{sec:dense}

In \Cref{sec:intro,sec:method}, we hypothesized that a shallow loop of a looped model is a strong multi-token drafter because all-depth supervision (\ref{eq:loop_loss}) encourages shallow states to encode future tokens while sharing all model parameters, and because conditioning on the deep prefix cache ${\bf c}^{<t}_R$ provides full-depth context representations. We now test both factors through controlled comparisons.

\paragraph{All-depth supervision and full-parameter sharing.}
We compare a single loop iteration of \ours ($E=0$) against separate draft networks and early layers of non-looped models trained on the same data with the same objective (\ref{eq:loss}), evaluated in a shared PyTorch implementation at batch size one.

\begin{wraptable}{r}{0.50\linewidth}\vspace{-1.2em}
\captionof{table}{\textbf{\ours gives larger speedups than a separate drafter.} Speedup ($\times$) over plain decoding, with accepted tokens per step in parentheses. Params.\ is the number of trained parameters, and depth is the number of layer evaluations in one drafting pass relative to a full decoding step.}
\label{tab:dense_controls}
\centering
\footnotesize
\setlength{\tabcolsep}{2.5pt}
\begin{tabular}{lrccc}
\toprule
Drafter & Params. & Depth & GSM8K & MATH-500 \\
\midrule
Separate, 1 layer & 56M & $1\%$ & 2.55 (2.9) & 2.85 (2.9) \\
Separate, 4 layers & 210M & $4\%$ & 3.70 (4.5) & 4.09 (4.4) \\
Separate, DFlash & 278M & $5\%$ & \underline{4.32} (5.1) & \underline{4.05} (4.6) \\
\rowcolor{LoopedFlowLight} \ours & 61M & $25\%$ & \textbf{5.48} (7.6) & \textbf{5.54} (7.0) \\
\bottomrule
\end{tabular}
\vspace{-0.8em}
\end{wraptable}

First, on Ouro~1.4B (\Cref{tab:dense_controls}), a single loop of \ours runs $25\%$ of the layers of a decoding step, yet its higher acceptance ($7.0$--$7.6$ tokens per step) yields a $5.5\times$ speedup on both GSM8K and MATH-500. This outperforms the 5-layer DFlash block drafter~\citep{chen2026dflash}, which reaches $4.3\times$ and $4.1\times$ despite having $4.5\times$ more trained parameters than our adapters ($278$M vs.\ $61$M), as well as 1-layer and 4-layer EAGLE-style separate drafters~\citep{li2024eagle} at $2.6$--$4.1\times$. Even though a separate drafter runs fewer layers per pass, its small network has limited capacity and must learn representations from scratch, whereas adapting a single loop leverages all pretrained parameters of the target model to achieve a much higher acceptance with far fewer trained parameters.

\begin{wrapfigure}{r}{0.46\linewidth}
 \vspace{-1.0em}
 \centering
 \includegraphics[width=\linewidth]{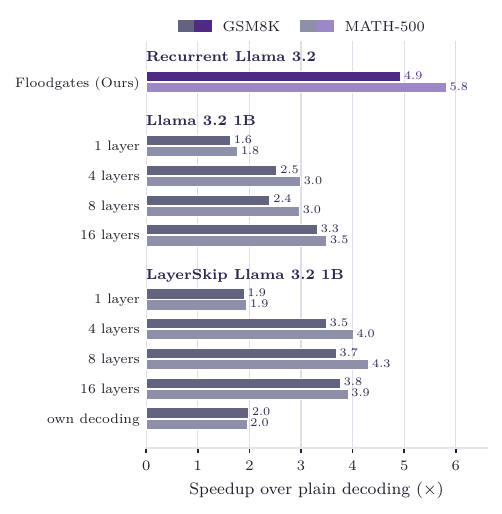}
 \vspace{-1.4em}
\caption{\textbf{\ours gives larger speedups than any number of layers of a non-looped model.} Speedup over plain decoding on GSM8K and MATH-500. On Llama 3.2 1B and LayerSkip Llama 3.2 1B, we train our drafter on the first $1$ to $16$ layers, and ``own decoding'' is the native self-speculative decoding of LayerSkip.}

 \label{fig:drafters}
 \vspace{-0.8em}
 \end{wrapfigure}

 Second, to separate the effect of early-depth supervision from that of looped parameter sharing, we compare Recurrent~Llama~3.2 against the 16-layer non-looped model it was built from, Llama 3.2 1B, and against LayerSkip Llama 3.2 1B~\citep{elhoushi2024layerskip}, which finetunes Llama 3.2 1B with next-token losses at early layers (\Cref{fig:drafters}). On both non-looped models, we apply our method to the first $L_d\in\{1,\dots,16\}$ layers by training LoRA adapters on those $L_d$ layers to draft $k$ mask tokens in parallel, which are then verified by all $16$ layers.
By comparing Llama 3.2 1B and LayerSkip, we isolate the benefit of supervising early depths. When drafting with only the first layer ($L_d=1$, which matches the $7\%$ relative depth of one loop of Recurrent~Llama~3.2), Llama 3.2 1B reaches only $1.6$--$1.8\times$ speedup on GSM8K and MATH-500, whereas early-exit supervision in LayerSkip raises the speedup of our drafter to $1.9\times$ on the first layer and from $2.5$--$3.0\times$ to $3.5$--$4.0\times$ on the first four layers ($L_d=4$). Consistent with this gain, training linear probes~\citep{pal2023future} on each layer of both non-looped models to predict the next three tokens shows that early-exit supervision improves accuracy at all three future positions across all five benchmarks (\Cref{sec:app_probe}, \Cref{fig:motivation_full}). Even so, the non-looped drafters peak at $3.5$--$4.3\times$ with half or all $16$ layers, whereas a single loop of Recurrent~Llama~3.2 reaches $4.9\times$ and $5.8\times$ while running just $7\%$ of the layers of a decoding step. This is because the first $L_d$ layers of a non-looped model use only $L_d/16$ of the parameters and attend only to their own shallow prefix states, whereas one loop of Recurrent~Llama~3.2 uses all parameters of the shared block and attends to the deep prefix cache ${\bf c}^{<t}_R$.

\begin{figure}[t]
\sbox0{\includegraphics[width=0.57\linewidth]{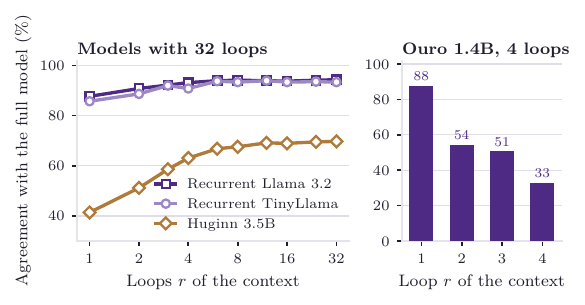}}
\sbox1{\small\setlength{\tabcolsep}{3pt}\renewcommand{\arraystretch}{1.1}\begin{tabular}{lcc}
\toprule
Benchmark & Final loop ${\bf c}^{<t}_R$ & First loop ${\bf c}^{<t}_1$ \\
\midrule
\multicolumn{3}{l}{\textit{Recurrent Llama 3.2}} \\
GSM8K & \textbf{4.93} (5.9) & 4.35 (5.2) \\
MATH-500 & \textbf{6.47} (7.8) & 5.89 (7.1) \\
HumanEval+ & \textbf{5.37} (6.5) & 4.61 (5.5) \\
MBPP+ & \textbf{4.53} (5.4) & 4.01 (4.8) \\
BBH & \textbf{4.47} (5.4) & 3.39 (4.1) \\
\midrule
\multicolumn{3}{l}{\textit{Ouro 1.4B}} \\
GSM8K & \textbf{5.04} (7.5) & 4.42 (6.6) \\
MATH-500 & \textbf{4.69} (7.0) & 4.05 (6.1) \\
HumanEval+ & \textbf{4.83} (7.2) & 4.33 (6.5) \\
MBPP+ & \textbf{3.01} (4.6) & 2.70 (4.1) \\
BBH & \textbf{4.40} (6.6) & 3.62 (5.4) \\
\bottomrule
\end{tabular}}
\dimen0=\dimexpr\ht0+\dp0\relax \ifdim\dimexpr\ht1+\dp1\relax>\dimen0 \dimen0=\dimexpr\ht1+\dp1\relax \fi   %
\begin{minipage}[t]{0.57\linewidth}
\centering
\begin{minipage}[b][\dimen0][c]{\linewidth}\centering\usebox0\end{minipage}
\captionof{figure}{\textbf{Without training, a deeper prefix cache helps one loop on the 32-loop models but not on Ouro~1.4B.} Next-token agreement on GSM8K between the full model and one untrained loop that attends to the prefix cache after $r$ loops.}
\label{fig:mature}
\end{minipage}\hfill
\begin{minipage}[t]{0.40\linewidth}
\centering
\begin{minipage}[b][\dimen0][c]{\linewidth}\centering\usebox1\end{minipage}
\captionof{table}{\textbf{After training, the final-loop cache gives larger speedups on both models.} Speedup ($\times$) over plain decoding with a single pass ($E=0$), with $\tauacc$ in parentheses. The two drafters of each model differ only in the cache they read.}

\label{tab:context}
\end{minipage}
\end{figure}

\paragraph{Conditioning on the deep prefix cache.}
During pretraining, each loop attends to the context from the same loop depth, so the first loop only sees the shallowest states of earlier tokens (${\bf c}^{<t}_1$) and never sees the final-loop cache (${\bf c}^{<t}_R$). To see how cache depth affects the first loop before any training, we run one untrained loop over the current token with the prefix cache after $r$ loops and measure how often its predicted next token agrees with the full model (\Cref{fig:mature}). On the 32-loop models, moving from the first-loop cache ($r=1$) to the final-loop cache ($r=32$) raises agreement from $86$--$88\%$ to $93$--$94\%$ on Recurrent~Llama~3.2 and Recurrent~TinyLlama, and from $41\%$ to $70\%$ on Huginn 3.5B. On Ouro~1.4B, however, agreement drops from $88\%$ at the first-loop cache ($r=1$) to $33\%$ at the final-loop cache ($r=4$), because Ouro loops its entire layer stack and its first loop cannot directly read the final-loop cache without adaptation. To test whether the deep prefix cache is still better once the drafter is trained, we train drafters under identical settings conditioned on either ${\bf c}^{<t}_R$ or ${\bf c}^{<t}_1$ (\Cref{tab:context}). Once adapted, Ouro~1.4B reverses the untrained trend and accepts $4.6$--$7.5$ tokens per step from ${\bf c}^{<t}_R$ compared to $4.1$--$6.6$ from ${\bf c}^{<t}_1$, decoding $11$--$22\%$ faster across all five benchmarks, while Recurrent~Llama~3.2 decodes $10$--$32\%$ faster with ${\bf c}^{<t}_R$ than with ${\bf c}^{<t}_1$. Together, these results confirm that both the deep prefix cache and adapter training are essential for fast drafting.

\subsection{Elastic drafting across verifier depths (Q3)}
\label{sec:controls}

\begin{figure}[t]
\centering
\includegraphics[width=\linewidth]{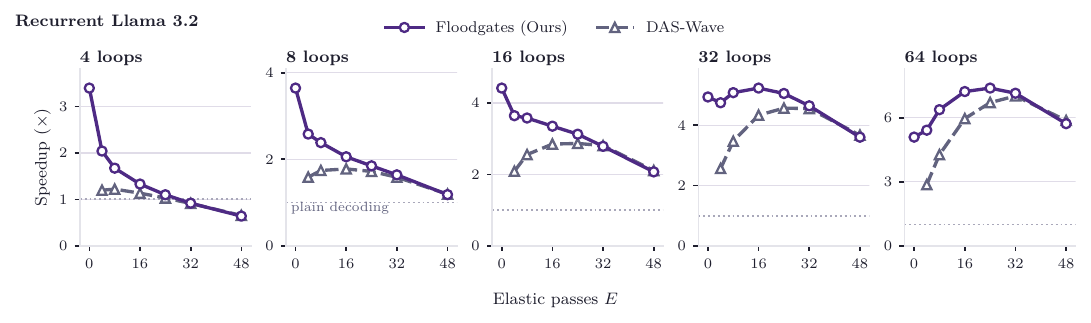}
\caption{\textbf{Elastic passes increase the speedup only on deep verifiers.} Speedup over plain decoding on GSM8K against the elastic drafting budget $E$. Recurrent~Llama~3.2 runs at $4$ to $64$ loops with a single drafter trained at $32$ loops. Up to $16$ loops, a single pass ($E=0$) is fastest, whereas at $32$ and $64$ loops the speedup peaks at $16$ to $24$ passes. DAS-Wave runs the same loops starting from copies of the last token.}

\label{fig:budget}
\end{figure}

\begin{figure}[t]
\centering
\includegraphics[width=\linewidth]{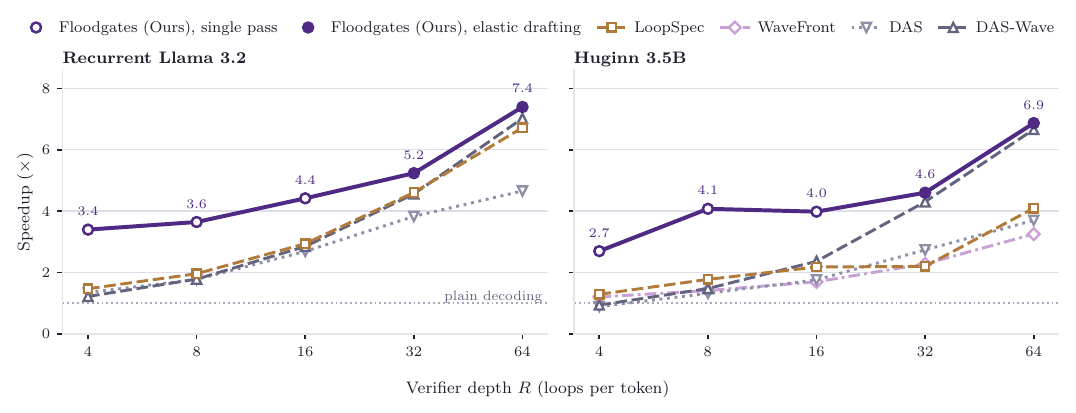}
\caption{\textbf{\ours is the fastest method at every verifier depth.} Speedup over plain decoding on GSM8K when Recurrent~Llama~3.2 and Huginn 3.5B run at $4$ to $64$ loops. Each method uses its best setting, and \ours uses a single drafter per model trained at $32$ loops. At each depth, we select the elastic drafting budget $E$ on held-out problems. Open circles mark the depths where a single pass ($E=0$) is selected. Filled circles mark the depths where elastic passes ($E>0$) are preferred (the deeper depths).}
\label{fig:depth}
\end{figure}

A key advantage of looped models is that their compute can be scaled at test time by varying the number of loops $R$ without retraining~\citep{geiping2025scaling,suleymanzade2026thinking}. In \Cref{sec:refine}, we showed that because each elastic pass of \ours costs a single loop iteration ($1/R$ of a decoding step) and requires no extra parameters, a single trained drafter can adjust its elastic drafting budget $E$ to the depth of the verifier. We now test how the optimal budget $E$ changes with verifier depth and how \ours compares with prior methods across test-time depths.

\paragraph{Elastic drafting budget.}
\Cref{fig:budget} shows how speedup varies with the number of elastic passes $E$ as the verifier depth of Recurrent~Llama~3.2 scales from $4$ to $64$ loops, using a single drafter trained at $32$ loops. On shallower verifiers running up to $16$ loops, where each loop is a large fraction of a decoding step, single-pass drafting ($E=0$) is fastest and additional passes reduce speedup, whereas for deeper verifiers running $32$ and $64$ loops, the speedup rises up to $16$--$24$ elastic passes before declining. At every depth and budget $E\le 24$, \ours outperforms DAS-Wave, which runs the same loops starting from copies of the last token. The gap is largest when only a few elastic passes are run ($63$--$89\%$ higher speedup at $4$ passes, $37$--$49\%$ at $8$ passes, and $16$--$21\%$ at $16$ passes), because \ours starts its elastic passes from a trained multi-token draft (\Cref{sec:app_positions} shows per-position acceptance rates across $E$).

\paragraph{Effect of model depth.}
Choosing the elastic drafting budget $E$ by optimizing on held-out problems at each depth, we compare \ours against all baselines as the test-time depth of Recurrent~Llama~3.2 and Huginn 3.5B varies from $4$ to $64$ loops (\Cref{fig:depth}). \ours is the fastest method at every depth on both models. From $4$ to $16$ loops, where single-pass drafting ($E=0$) is selected, \ours already achieves $3.4$--$4.4\times$ speedup on Recurrent~Llama~3.2 and $2.7$--$4.1\times$ on Huginn 3.5B. At $32$ and $64$ loops, adding elastic passes raises the speedup to $5.2\times$ and $7.4\times$ on Recurrent~Llama~3.2 and to $4.6\times$ and $6.9\times$ on Huginn 3.5B. In contrast, training-free methods use untrained loops that predict only one new token per step, making them slow on shallow models ($1.3$--$1.5\times$ for LoopSpec, $0.9$--$1.4\times$ for DAS, and $0.9$--$1.2\times$ for DAS-Wave at $4$ loops) and competitive only at $32$ loops and beyond.

\subsection{High-throughput batched serving (Q4)}
\label{sec:batch}

\begin{figure}[t]
\centering
\includegraphics[width=\linewidth]{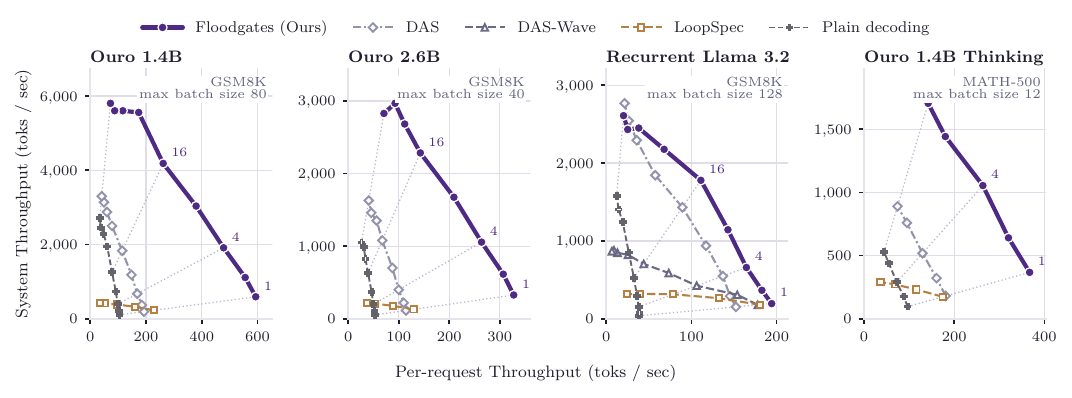}
\caption{\textbf{\ours has the highest serving throughput at every batch size on the Ouro models, and up to batch size 64 on Recurrent~Llama~3.2.} System throughput (tokens/s over all requests) against per-request throughput on an A100 under continuous batching. Each point is one batch size, which grows from $1$ at the bottom right to the largest that fits in memory at the top left. }
\label{fig:serving}
\end{figure}

In \Cref{sec:method_batch}, we argued that \ours is well suited to concurrent serving because it avoids depth pipelining and reads only the final-loop prefix cache ${\bf c}^{<t}_R$ once per draft block, while dynamically adapting the number of verified drafts $M^\star$ (\ref{eq:batch_select}) as the batch size grows. To test whether these properties translate to high-throughput serving under hardware saturation, we evaluate continuous batching across varying batch sizes.

\paragraph{Setup.}
Following \citet{sahoo2026uno}, we evaluate continuous batching on an A100-80GB GPU across batch sizes up to the device memory limit, decoding all 1,319 GSM8K test problems with Ouro~1.4B, Ouro~2.6B, and Recurrent~Llama~3.2, and all 500 MATH-500 problems with Ouro~1.4B~Thinking under greedy decoding. At each batch size, we report total system throughput (tokens/s across all requests) and per-request throughput (system throughput divided by batch size). Since LoopSpec's official implementation decodes one request at a time, we run as many instances of its server on the GPU as fit in memory.

\paragraph{Results.}
\Cref{fig:serving} shows the results. On Ouro~1.4B and Ouro~2.6B, \ours achieves the highest system throughput at every batch size. At the largest batch that fits in memory, it generates $2.1\times$ and $2.7\times$ as many tokens per second as plain decoding ($5{,}805$ and $2{,}827$ tokens/s), outperforming the strongest baseline, DAS, which reaches $1.2\times$ and $1.5\times$, while replicated LoopSpec servers exhaust memory early and peak at $0.15$--$0.21\times$. On Recurrent~Llama~3.2, \ours leads up to batch size $64$, while DAS slightly leads at the two largest batch sizes ($1.75\times$ vs.\ $1.65\times$ at batch size $128$). Across these models, our adaptive verification budget $M^\star$ (\ref{eq:batch_select}) matches fixed-block verification (which verifies all $k$ drafts for every request) at small batch sizes and increases throughput at the largest batch size by $19\%$ on Ouro~1.4B and $54\%$ on Recurrent~Llama~3.2 (\Cref{sec:app_serving_results}). On Ouro~1.4B~Thinking, where long reasoning chains limit the maximum batch size to $12$, \ours reaches $3.2\times$ the throughput of plain decoding and $1.9\times$ that of DAS.

Furthermore, because \ours does not batch tokens at different loop depths as pipelined methods do, it can be combined directly with cross-loop cache sharing~\citep{prairie2026benice}, where all $R$ verification loops reuse only the final-loop key-value cache. In \Cref{sec:app_share} (\Cref{tab:share_acc,fig:serving_huginn}), we show that serving Huginn 3.5B with a single shared cache across loops preserves accuracy within $2$ points while scaling the maximum batch size on GSM8K from $16$ to $320$, where \ours achieves the highest throughput at every batch size across all five benchmarks ($1.08$--$1.33\times$ over the strongest baseline at the largest batch size).

\section{Related work}
\label{sec:related}
\paragraph{Speculative decoding and drafter capacity.}
Speculative decoding accelerates generation without changing the target distribution by using a fast draft model to propose candidate tokens and verifying them in a single forward pass of the target model~\citep{leviathan2023fast,chen2023accelerating}. To achieve a large speedup, the drafter must be both much cheaper to run than the target model and accurate enough that most of its proposals are accepted~\citep{leviathan2023fast,zhou2024distillspec}. In dense models, where every layer is applied once per token, these two requirements conflict. Fast drafters are therefore either small separate networks trained on target features~\citep{li2024eagle,li2024eagle2,li2025eagle3,chen2026dflash,hui2026peagle,cheng2026dspark,huang2026domino,chen2026dloop}, or self-speculative methods that skip or exit early from layers of the target model~\citep{zhang2024draft,elhoushi2024layerskip,liu2024kangaroo,xia2025swift,lin2026seed}. In both cases, the drafter uses only a fraction of the parameters of the target model, which limits its verification rate. On looped models, \citet{cho2026loopspec} even found a separate drafter slower than training-free self-speculation. \ours avoids this tradeoff by using a shallow loop of the model's own shared block as the drafter, reducing computation while keeping all parameters of the model.

\paragraph{Multi-token prediction and mask-token drafting.}
To propose multiple tokens without running the drafter sequentially, a model can be trained to predict several future tokens in parallel. One line of work adds auxiliary output heads that predict future tokens from the final hidden state~\citep{cai2024medusa,ankner2024hydra,gloeckle2024better,deepseekai2024deepseekv3}, or pretrains a looped model so that later loops predict later tokens~\citep{shomali2026loopmtp}. A second line of work, which is closest to our drafting pass, appends placeholder or mask tokens as a suffix and adapts the model with a small number of trained parameters to fill them in at once~\citep{monea2023pass,bhendawade2024streaming,samragh2025your,sahoo2026uno}. On a dense model, however, the mask tokens pass through all layers, so a separate drafting pass costs about as much as verification, which can cancel the speedup~\citep{javat2026accepted}. In a looped model, a drafting pass can be as cheap as a single loop iteration, so \ours drafts from the accepted tokens at a small fraction of a decoding step.

\paragraph{Looped language models and their decoding.}
Looped language models apply a shared block of layers repeatedly to every token~\citep{dehghani2019universal,geiping2025scaling,zhu2025scaling,mcleish2025teaching,bae2025relaxed}, and the number of loops can be chosen at inference time~\citep{yu2026tloopformer,schwethelm2026depth}. Prior methods for fast decoding of looped models use the loop without training. LoopSpec and WaveFront take the prediction after an early loop as the draft for the next token \citep{cho2026loopspec, ha2026wavefront}, building on pipelined speculative execution \citep{hooper2023speed}. Because the model has only been trained to predict the immediate next token, this yields at most one draft token per loop, and because later tokens start their loops before earlier tokens finish, it requires storing a separate key-value cache for every loop depth. DAS and DAS-Wave instead let shallow passes attend to the final-loop prefix cache \citep{li2026shallow}, but because the loop is not trained to look ahead, DAS drafts one token at a time and DAS-Wave refines a window of drafts over many shallow passes, which pays off only on models with many loops. Similarly, the parallel sampler of \citet{geiping2026efficient} decodes multiple tokens in parallel without a verification step, which changes the outputs of the model and thus is not lossless. \ours shares with DAS-Wave the deep prefix cache and the use of the model's own loops for elastic passes, but starts from a trained multi-token draft, so that a single loop iteration already yields several tokens and elastic passes are used only when a single loop is cheap.

\section{Conclusion}
\label{sec:conclusion}

We presented \ours, a simple self-speculative decoding method that adapts a shallow loop of a looped language model to draft several upcoming tokens in parallel over suffix mask tokens. Across seven looped models and five benchmarks, \ours outperforms prior lossless decoding methods at batch size one, and its speedups carry over to high-throughput batched serving. Our analysis shows that a shallow loop makes an effective drafter because all-depth training already encourages early states to carry information about future tokens, and because conditioning on the deep prefix cache lets a single loop build on full-depth context computation. These findings suggest that looped models are naturally suited for fast multi-token generation once their shallow loops are trained to look ahead.

\bibliographystyle{plainnat}
\bibliography{references}

\beginappendix
\crefalias{section}{appendix}
\crefalias{subsection}{appendix}

\renewcommand{\topfraction}{0.9}
\renewcommand{\bottomfraction}{0.6}
\renewcommand{\textfraction}{0.07}
\renewcommand{\floatpagefraction}{0.8}
\newpage

\section{Experimental details}
\label{sec:app_details}We describe the evaluated models and benchmarks in \Cref{sec:app_models}, drafter training in \Cref{sec:app_training}, our inference engine in \Cref{sec:app_engine}, and baseline configurations in \Cref{sec:app_baselines}.
\subsection{Models and tasks}
\label{sec:app_models}\paragraph{Models.}
\Cref{tab:models} summarizes the seven looped models evaluated in our experiments. Ouro~1.4B and Ouro~2.6B~\citep{zhu2025scaling}, along with their reasoning variants Ouro~1.4B~Thinking and Ouro~2.6B~Thinking, loop their entire layer stack $R=4$ times. Recurrent~Llama~3.2, Recurrent~TinyLlama~\citep{mcleish2025teaching}, and Huginn 3.5B~\citep{geiping2025scaling} run an initial prelude once, loop a middle core block, and run a final coda once. Because these three models were trained with loop counts sampled randomly, they can be evaluated at arbitrary loop depths at inference time. Recurrent~Llama~3.2 and Recurrent~TinyLlama were adapted from pretrained Llama 3.2 1B and TinyLlama 1.1B checkpoints~\citep{mcleish2025teaching}, while Huginn 3.5B was pretrained from scratch~\citep{geiping2025scaling}. All models are evaluated using their public checkpoints in bfloat16 precision.\begin{table}[!htb]
\centering
\caption{\textbf{The Ouro models loop all of their layers four times, and the other three models loop a middle block 32 times.} Block denotes the looped layers, $R$ the default number of loops, and layers per token the total number of layer evaluations in one decoding step.}
\label{tab:models}
\small
\setlength{\tabcolsep}{5pt}
\begin{tabular}{llccc}
\toprule
Model & Block & $R$ & Layers per token & Width \\
\midrule
Ouro~1.4B, Ouro~1.4B~Thinking & entire stack, $24$ layers & $4$ & $96$ & $2048$ \\
Ouro~2.6B, Ouro~2.6B~Thinking & entire stack, $48$ layers & $4$ & $192$ & $2048$ \\
Recurrent~Llama~3.2 & prelude $4$, core $6$, coda $4$ & $32$ & $200$ & $2048$ \\
Recurrent~TinyLlama & prelude $4$, core $8$, coda $4$ & $32$ & $264$ & $2048$ \\
Huginn 3.5B & prelude $2$, core $4$, coda $2$ & $32$ & $132$ & $5280$ \\
\bottomrule
\end{tabular}
\end{table}\paragraph{Tasks.}
\Cref{tab:tasks} lists the evaluation benchmarks, comprising GSM8K~\citep{cobbe2021training}, MATH-500~\citep{lightman2024lets}, HumanEval+ and MBPP+~\citep{chen2021evaluating,austin2021program,liu2023evalplus}, and BBH~\citep{suzgun2023challenging}. Following LoopSpec~\citep{cho2026loopspec}, we use the prompts, few-shot formats, generation limits, and stop strings implemented in \texttt{lm-evaluation-harness}~\citep{gao2024evalharness}. Prompts on GSM8K and BBH use 3-shot chain-of-thought examples, and prompts on MATH-500 use 4-shot examples. On HumanEval+ and MBPP+, each prompt provides a Python function signature and docstring, and the model completes the function body. The reasoning models use their native chat template with thinking enabled, capped at 4,096 new tokens on GSM8K and 8,192 on MATH-500. A generation is scored as correct on GSM8K and BBH if the extracted final answer matches the reference, on MATH-500 if it is mathematically equivalent to the reference, and on HumanEval+ and MBPP+ if the completed function passes all test cases. Recurrent~TinyLlama has a 2,048-token context limit, which excludes BBH prompts and caps its MATH-500 generation length at 1,024 tokens.\begin{table}[!htb]
\centering
\caption{\textbf{All benchmarks follow the evaluation protocol of LoopSpec.} We list the number of test problems, few-shot examples, maximum new tokens, and stop strings of each benchmark~\citep{cho2026loopspec}.}
\label{tab:tasks}
\small
\setlength{\tabcolsep}{5pt}
\begin{tabular}{llccl}
\toprule
Task & Problems & Shots & Max.\ new tokens & Stop strings \\
\midrule
GSM8K & first $100$ of $1{,}319$ & $3$ & $512$ & \texttt{Q:}, end of text \\
MATH-500 & first $100$ of $500$ & $4$ & $2{,}048$ & \texttt{Problem:} \\
HumanEval+ & all $164$ & $0$ & $1{,}024$ & end of function \\
MBPP+ & all $378$ & $0$ & $1{,}024$ & end of function \\
BBH & $20$ per task, $540$ & $3$ & $1{,}024$ & \texttt{Q}, blank line, end of text \\
\bottomrule
\end{tabular}
\end{table}
\subsection{Training details}
\label{sec:app_training}\paragraph{Data.}
We train each drafter on greedy generations from its target model. For the five base models, we generate continuations on 7,473 GSM8K training problems, 7,500 MATH training problems~\citep{hendrycks2021measuring}, 3,000 Python function signatures and docstrings from CodeSearchNet~\citep{husain2019codesearchnet}, and $256$-token prefixes from 4,000 FineWeb-Edu documents~\citep{penedo2024fineweb}. GSM8K and MATH prompts follow the 3-shot and 4-shot evaluation formats, and CodeSearchNet prompts follow the HumanEval+/MBPP+ format. For the reasoning models, we generate continuations on the 7,473 GSM8K training problems and 2,032 MATH training problems using their chat template with thinking enabled and a limit of 4,096 new tokens.\paragraph{Adapters and loss.}
We apply rank-$64$ LoRA adapters~\citep{hu2022lora} to all attention and feed-forward projections. On Recurrent~Llama~3.2, Recurrent~TinyLlama, and Huginn 3.5B, adapters are placed on the prelude, core, and coda layers, as well as the input projection that combines the token embedding with the recurrent state. Adapters and the mask embedding ${\bf m}$ are initialized to zero. In the training loss (\ref{eq:loss}), we set slot weights $w_j \propto e^{-(j-1)/4}$ normalized to sum to one. On the Ouro models, each mask slot predicts the token at its own position, producing $k$ drafts from $k$ mask tokens. On the other three models, each position predicts the next token, so the slot of the last verified token ${\bf x}^t$ also produces a draft, and $k$ mask tokens yield $k+1$ drafts.\paragraph{Optimization.}
We train with AdamW~\citep{loshchilov2019decoupled} using decay rates $(\beta_1, \beta_2)=(0.9, 0.95)$ for the first and second moments, zero weight decay, a peak learning rate of $4\times10^{-4}$ with $100$ warmup steps and linear decay to $4\times10^{-5}$, gradient clipping at $1.0$, and a batch size of $8$ sequences. For each training sequence, we sample $64$ positions uniformly from the generated continuation and append a block of $k$ mask slots after each position. A single forward pass of the frozen target model computes the deep prefix cache ${\bf c}^{<t}_R$ for all sampled positions in a sequence. Attention masks prevent slots placed at different sampled positions from attending to one another. The frozen model runs in bfloat16 and the adapters in float32. \Cref{tab:hparams} lists the hyperparameters and training time for each model, and \Cref{sec:app_design} ablates the main design choices.\begin{table}[!htb]
\centering
\caption{\textbf{Each drafter trains 31M to 121M parameters in 4 to 15 hours on one GPU.} Drafts per pass is the number of tokens drafted by one loop, and $E$ is the elastic drafting budget selected on held-out problems (\Cref{sec:app_passes}).}
\label{tab:hparams}
\scriptsize
\setlength{\tabcolsep}{3pt}
\begin{tabular}{lccccccc}
\toprule
& Ouro~1.4B & Ouro~2.6B & 1.4B~Thinking & 2.6B~Thinking & Rec.\ Llama & Rec.\ TinyLlama & Huginn \\
\midrule
Training sequences & $21{,}973$ & $21{,}973$ & $9{,}505$ & $9{,}505$ & $21{,}973$ & $21{,}973$ & $21{,}973$ \\
Mask tokens $k$ & $12$ & $12$ & $8$ & $8$ & $12$ & $12$ & $12$ \\
Drafts per pass & $12$ & $12$ & $8$ & $8$ & $13$ & $13$ & $13$ \\
Adapter rank & $64$ & $64$ & $64$ & $64$ & $64$ & $64$ & $64$ \\
Trained parameters & $60.6$M & $121.1$M & $60.6$M & $121.1$M & $34.3$M & $30.8$M & $50.2$M \\
Training steps & $32$K & $32$K & $16$K & $16$K & $32$K & $32$K & $32$K \\
Elastic passes $E$ & $0$ & $0$ & $0$ & $0$ & $24$ & $24$ & $24$ \\
Training time & $6.3$ h & $9.2$ h & $3.8$ h & $7.3$ h & $8.4$ h & $8.9$ h & $14.7$ h \\
GPU & H200 & B300 & B300 & B300 & H100 & H100 & B300 \\
\bottomrule
\end{tabular}
\end{table}
\subsection{Implementation details}
\label{sec:app_algo}\label{sec:app_step}\label{sec:app_lossless}\label{sec:app_engine}\label{sec:app_batch}We implement our decoding engine in PyTorch with static key-value buffers for each loop depth, CUDA graphs, and Triton kernels. The training-free baselines in \Cref{tab:main} (except LoopSpec and WaveFront, which use the authors' code) run in this engine and share the same verification step. Plain decoding in our engine is $32\%$ faster on Ouro~1.4B than in SGLang, ensuring that our speedups are measured against a strong baseline. All inference experiments are run on a single NVIDIA A100-80GB GPU in bfloat16.
\FloatBarrier
\subsection{Baseline settings}
\label{sec:app_baselines}\paragraph{LoopSpec and WaveFront.}
We use the official implementations from the authors. LoopSpec serves one request at a time in SGLang~\citep{zheng2024sglang} and is benchmarked against SGLang's plain decoding. Following the authors' settings, we use proposal depths $d_1=1$ for one proposal (LoopSpec$^\ast$) and $(d_1,d_2)=(1,2)$ for the full method on the Ouro models, and $d_1=2$ and $(2,8)$ on the 32-loop models. WaveFront is evaluated in its lossless setting on the models it supports, drafting a token after every loop on the Ouro models and after every $T_d$ loops on Huginn 3.5B, where we report the better of $T_d=4$ and $T_d=8$.\paragraph{DAS and DAS-Wave.}
Since neither method has public code, we reimplemented both following \citet{li2026shallow} in our engine with the same verification step as \ours. DAS drafts eight tokens sequentially, where each draft position runs $r$ loops conditioned on the deep prefix cache ($r=2$ on Recurrent~Llama~3.2 and Recurrent~TinyLlama, and $r=4$ on Huginn 3.5B). On the Ouro models, where an untrained first loop cannot read the final-loop cache (\Cref{fig:mature}), DAS runs one loop per draft conditioned on the first-loop cache, which gives its best results. DAS-Wave updates a window of up to $\gamma$ positions in parallel over $J$ refinement rounds. Each round runs $k$ loops over the window conditioned on the deep prefix cache, where each position takes as input the token predicted at the preceding position in the previous round and the window grows by one position per round. On the 32-loop models, we use the paper's settings ($\gamma=32$ and $J=24$, with $k=1$ on Recurrent~Llama~3.2 and Recurrent~TinyLlama and $k=2$ on Huginn 3.5B). On the 4-loop Ouro models, where each round costs $1/4$ of a decoding step, we use $J=12$ rounds over $\gamma=16$ positions on the base models and $J=8$ rounds over $\gamma=32$ positions on the reasoning models, which performed best among the settings we tested.\paragraph{Jacobi decoding and prompt lookup decoding.}
Jacobi decoding iteratively refines a window of candidate tokens over multiple passes of the full model~\citep{santilli2023accelerating, fu2024break}. We use a window of $32$ positions with two model passes conditioned on the deep prefix cache on the 32-loop models, and report the best of one, two, or four passes on the Ouro models. Prompt lookup decoding searches the prefix for the most recent match of the last $n$ tokens and proposes the $\gamma$ tokens that followed it~\citep{saxena2023pld}. We report the better of $(n,\gamma)=(2,8)$ and $(3,12)$.

\paragraph{Trained baselines.}
The separate drafters in \Cref{sec:dense} are trained for Ouro~1.4B on the same four-domain data and for the same $32$K steps as \ours (\Cref{sec:app_training}). The 1-layer and 4-layer separate drafters follow EAGLE~\citep{li2024eagle}, where each position combines its token embedding with the final hidden state of Ouro~1.4B, passes through the drafter layers, and is read out by the frozen head of Ouro~1.4B over $k=12$ mask tokens using the loss in (\ref{eq:loss}). The DFlash drafter~\citep{chen2026dflash}, following LoopSpec~\citep{cho2026loopspec}, uses five layers at the width of Ouro~1.4B that attend to hidden states from five layers of Ouro~1.4B and drafts a block of $16$ tokens from mask tokens, trained with the optimizer settings and loss weights of DFlash. On Llama 3.2 1B and LayerSkip Llama 3.2 1B~\citep{elhoushi2024layerskip}, we replace the loop with the first $L_d$ layers of the dense model and train LoRA adapters on those layers over $k=12$ mask tokens using each model's own generations (\Cref{sec:app_training}). For LayerSkip's native self-speculative decoding, we draft one token at a time from the exit after layer $L_d$, verify with the remaining layers, and report the speedup at its best $L_d$.
\section{Additional results}
\label{sec:app_more}
\subsection{Single-pass vs.\ elastic drafting}
\label{sec:app_passes}
The elastic drafting budget $E$ is the only inference hyperparameter of \ours. For each model, we select $E$ on 100 held-out problems (test problems 101--150 of GSM8K and MATH-500, excluded from \Cref{tab:main,tab:thinking}). We evaluate $E\in\{0,1,2\}$ on the Ouro models and $E\in\{0,8,16,24,32\}$ on the 32-loop models, choosing the $E$ with the highest mean speedup over the two tasks. This selects $E=0$ on the four Ouro models and $E=24$ on the three 32-loop models. \Cref{tab:passes} compares both settings across all models and tasks. On the Ouro models ($R=4$), where each elastic pass costs $1/4$ of a decoding step, a single pass ($E=0$) is faster on every task under both greedy decoding and sampling. On the 32-loop models ($R=32$), where each pass costs $1/32$ of a decoding step, elastic drafting ($E=24$) is faster on $12$ of the $14$ tasks under greedy decoding (except GSM8K and MATH-500 on Huginn 3.5B, where $E=0$ is faster by $0.4\%$ and $8\%$) and on $12$ of the $14$ tasks under sampling (except GSM8K on Recurrent~Llama~3.2 and Huginn 3.5B).\begin{table}[!htb]
\centering
\caption{\textbf{A single pass is fastest on the 4-loop Ouro models, and elastic drafting is faster on most tasks of the 32-loop models.} Speedup ($\times$) over plain decoding at batch size one, with tokens per step $\tauacc$ in parentheses, for a single pass ($E=0$) and for elastic drafting with $E$ passes. Shaded rows indicate the setting selected on held-out problems and reported in \Cref{tab:main,tab:thinking}. Reasoning models are sampled at temperature $0.6$ and top-$p$ $0.95$.}

\label{tab:passes}
\small
\setlength{\tabcolsep}{2.5pt}
\begin{tabular}{llccccc}
\toprule
Model & $E$ & GSM8K & MATH-500 & HumanEval+ & MBPP+ & BBH \\
\midrule
\multicolumn{7}{l}{\textit{Greedy decoding}} \\
\multirow{2}{*}{Ouro~1.4B} & \cellcolor{LoopedFlowLight}$0$ & \cellcolor{LoopedFlowLight}5.04 (7.5) & \cellcolor{LoopedFlowLight}4.69 (7.0) & \cellcolor{LoopedFlowLight}4.83 (7.2) & \cellcolor{LoopedFlowLight}3.01 (4.6) & \cellcolor{LoopedFlowLight}4.40 (6.6) \\
& $1$ & 3.75 (7.0) & 2.89 (5.4) & 3.37 (6.3) & 2.27 (4.4) & 3.29 (6.2) \\
\multirow{2}{*}{Ouro~2.6B} & \cellcolor{LoopedFlowLight}$0$ & \cellcolor{LoopedFlowLight}5.56 (8.2) & \cellcolor{LoopedFlowLight}6.04 (9.0) & \cellcolor{LoopedFlowLight}5.57 (8.3) & \cellcolor{LoopedFlowLight}3.39 (5.2) & \cellcolor{LoopedFlowLight}5.07 (7.5) \\
& $1$ & 3.43 (6.3) & 2.92 (5.4) & 3.14 (5.8) & 2.28 (4.4) & 3.15 (5.9) \\
\multirow{2}{*}{Ouro~1.4B~Thinking} & \cellcolor{LoopedFlowLight}$0$ & \cellcolor{LoopedFlowLight}3.70 (5.6) & \cellcolor{LoopedFlowLight}3.68 (5.5) & \cellcolor{LoopedFlowLight}-- & \cellcolor{LoopedFlowLight}-- & \cellcolor{LoopedFlowLight}-- \\
& $1$ & 2.56 (4.9) & 2.39 (4.5) & -- & -- & -- \\
\multirow{2}{*}{Ouro~2.6B~Thinking} & \cellcolor{LoopedFlowLight}$0$ & \cellcolor{LoopedFlowLight}3.90 (5.9) & \cellcolor{LoopedFlowLight}3.72 (5.5) & \cellcolor{LoopedFlowLight}-- & \cellcolor{LoopedFlowLight}-- & \cellcolor{LoopedFlowLight}-- \\
& $1$ & 2.19 (4.1) & 2.09 (3.9) & -- & -- & -- \\
\multirow{2}{*}{Recurrent~Llama~3.2} & $0$ & 4.93 (5.9) & 6.47 (7.8) & 5.37 (6.5) & 4.53 (5.4) & 4.47 (5.4) \\
& \cellcolor{LoopedFlowLight}$24$ & \cellcolor{LoopedFlowLight}5.04 (20.4) & \cellcolor{LoopedFlowLight}6.97 (28.5) & \cellcolor{LoopedFlowLight}6.91 (28.0) & \cellcolor{LoopedFlowLight}5.84 (23.5) & \cellcolor{LoopedFlowLight}6.62 (27.1) \\
\multirow{2}{*}{Huginn 3.5B} & $0$ & 4.50 (5.1) & 6.69 (7.5) & 4.77 (5.4) & 3.72 (4.2) & 3.66 (4.1) \\
& \cellcolor{LoopedFlowLight}$24$ & \cellcolor{LoopedFlowLight}4.48 (19.0) & \cellcolor{LoopedFlowLight}6.21 (26.6) & \cellcolor{LoopedFlowLight}6.18 (26.2) & \cellcolor{LoopedFlowLight}5.06 (21.4) & \cellcolor{LoopedFlowLight}5.70 (24.6) \\
\multirow{2}{*}{Recurrent~TinyLlama} & $0$ & 5.22 (6.2) & 6.07 (7.2) & 5.27 (6.2) & 4.72 (5.5) & -- \\
& \cellcolor{LoopedFlowLight}$24$ & \cellcolor{LoopedFlowLight}6.73 (22.2) & \cellcolor{LoopedFlowLight}7.79 (26.0) & \cellcolor{LoopedFlowLight}8.20 (27.0) & \cellcolor{LoopedFlowLight}7.17 (23.5) & \cellcolor{LoopedFlowLight}-- \\
\midrule
\multicolumn{7}{l}{\textit{Sampling at temperature $1$ and top-$p$ $0.7$}} \\
\multirow{2}{*}{Ouro~1.4B} & \cellcolor{LoopedFlowLight}$0$ & \cellcolor{LoopedFlowLight}4.16 (6.3) & \cellcolor{LoopedFlowLight}4.41 (6.7) & \cellcolor{LoopedFlowLight}4.49 (6.8) & \cellcolor{LoopedFlowLight}2.75 (4.3) & \cellcolor{LoopedFlowLight}4.11 (6.2) \\
& $1$ & 3.16 (6.0) & 2.63 (5.0) & 3.17 (6.0) & 2.09 (4.1) & 3.02 (5.7) \\
\multirow{2}{*}{Ouro~2.6B} & \cellcolor{LoopedFlowLight}$0$ & \cellcolor{LoopedFlowLight}4.50 (6.7) & \cellcolor{LoopedFlowLight}4.90 (7.3) & \cellcolor{LoopedFlowLight}4.82 (7.2) & \cellcolor{LoopedFlowLight}3.23 (4.9) & \cellcolor{LoopedFlowLight}4.77 (7.1) \\
& $1$ & 3.05 (5.7) & 2.57 (4.8) & 2.79 (5.2) & 2.15 (4.1) & 3.01 (5.6) \\
\multirow{2}{*}{Ouro~1.4B~Thinking} & \cellcolor{LoopedFlowLight}$0$ & \cellcolor{LoopedFlowLight}3.46 (5.2) & \cellcolor{LoopedFlowLight}3.23 (4.9) & \cellcolor{LoopedFlowLight}-- & \cellcolor{LoopedFlowLight}-- & \cellcolor{LoopedFlowLight}-- \\
& $1$ & 2.31 (4.5) & 2.14 (4.2) & -- & -- & -- \\
\multirow{2}{*}{Ouro~2.6B~Thinking} & \cellcolor{LoopedFlowLight}$0$ & \cellcolor{LoopedFlowLight}3.71 (5.5) & \cellcolor{LoopedFlowLight}3.41 (5.1) & \cellcolor{LoopedFlowLight}-- & \cellcolor{LoopedFlowLight}-- & \cellcolor{LoopedFlowLight}-- \\
& $1$ & 2.06 (3.9) & 1.97 (3.8) & -- & -- & -- \\
\multirow{2}{*}{Recurrent~Llama~3.2} & $0$ & 3.65 (4.4) & 4.47 (5.5) & 4.35 (5.3) & 4.23 (5.1) & 4.00 (4.9) \\
& \cellcolor{LoopedFlowLight}$24$ & \cellcolor{LoopedFlowLight}3.28 (7.6) & \cellcolor{LoopedFlowLight}4.66 (11.0) & \cellcolor{LoopedFlowLight}4.69 (10.9) & \cellcolor{LoopedFlowLight}4.34 (10.1) & \cellcolor{LoopedFlowLight}5.13 (12.1) \\
\multirow{2}{*}{Huginn 3.5B} & $0$ & 3.24 (3.7) & 4.53 (5.1) & 3.91 (4.4) & 2.93 (3.3) & 3.34 (3.8) \\
& \cellcolor{LoopedFlowLight}$24$ & \cellcolor{LoopedFlowLight}3.16 (7.3) & \cellcolor{LoopedFlowLight}4.60 (10.8) & \cellcolor{LoopedFlowLight}4.70 (10.9) & \cellcolor{LoopedFlowLight}3.52 (8.1) & \cellcolor{LoopedFlowLight}4.48 (19.3) \\
\multirow{2}{*}{Recurrent~TinyLlama} & $0$ & 3.73 (4.4) & 4.54 (5.4) & 4.16 (5.0) & 4.00 (4.8) & -- \\
& \cellcolor{LoopedFlowLight}$24$ & \cellcolor{LoopedFlowLight}4.01 (8.1) & \cellcolor{LoopedFlowLight}4.83 (9.9) & \cellcolor{LoopedFlowLight}5.61 (11.4) & \cellcolor{LoopedFlowLight}5.25 (10.6) & \cellcolor{LoopedFlowLight}-- \\
\bottomrule
\end{tabular}
\end{table}
\subsection{Confidence intervals}
\label{sec:app_ci}
\Cref{tab:ci} reports $95\%$ bootstrap confidence intervals for the greedy speedups in \Cref{tab:main}. For each task, we resample the test problems 2,000 times with replacement and compute the speedup of each method on the same resamples. The intervals remain within $5\%$ of the speedup for the training-free baselines and within $8\%$ for \ours.\begin{table}[!htb]
\centering
\caption{\textbf{The greedy speedups of \Cref{tab:main} vary by at most 8\% when the test problems are resampled.} Speedup ($\times$) over plain decoding with $95\%$ bootstrap intervals over problems, for \ours and the prior method with the highest mean speedup on each model.}
\label{tab:ci}
\scriptsize
\setlength{\tabcolsep}{1.5pt}
\begin{tabular}{llccccc}
\toprule
Model & Method & GSM8K & MATH-500 & HumanEval+ & MBPP+ & BBH \\
\midrule
\multirow{2}{*}{Ouro~1.4B} & LoopSpec & 2.77 [2.73, 2.80] & 2.81 [2.78, 2.84] & 3.10 [3.08, 3.11] & 2.83 [2.79, 2.88] & 2.86 [2.84, 2.88] \\
& \cellcolor{LoopedFlowLight}\ours & \cellcolor{LoopedFlowLight}\textbf{5.04} [4.84, 5.24] & \cellcolor{LoopedFlowLight}\textbf{4.69} [4.49, 4.91] & \cellcolor{LoopedFlowLight}\textbf{4.83} [4.69, 4.99] & \cellcolor{LoopedFlowLight}\textbf{3.01} [2.85, 3.25] & \cellcolor{LoopedFlowLight}\textbf{4.40} [4.28, 4.52] \\
\midrule
\multirow{2}{*}{Ouro~2.6B} & LoopSpec & 3.06 [3.03, 3.09] & 3.09 [3.02, 3.15] & 3.33 [3.31, 3.34] & 3.09 [3.07, 3.11] & 3.10 [3.08, 3.11] \\
& \cellcolor{LoopedFlowLight}\ours & \cellcolor{LoopedFlowLight}\textbf{5.56} [5.36, 5.77] & \cellcolor{LoopedFlowLight}\textbf{6.04} [5.59, 6.46] & \cellcolor{LoopedFlowLight}\textbf{5.57} [5.43, 5.70] & \cellcolor{LoopedFlowLight}\textbf{3.39} [3.30, 3.49] & \cellcolor{LoopedFlowLight}\textbf{5.07} [4.93, 5.20] \\
\midrule
\multirow{2}{*}{Rec.\ Llama 3.2} & LoopSpec & 4.61 [4.45, 4.77] & 6.11 [5.88, 6.28] & 6.03 [5.90, 6.14] & 5.41 [5.21, 5.60] & 5.65 [5.57, 5.73] \\
& \cellcolor{LoopedFlowLight}\ours & \cellcolor{LoopedFlowLight}\textbf{5.04} [4.75, 5.35] & \cellcolor{LoopedFlowLight}\textbf{6.97} [6.68, 7.16] & \cellcolor{LoopedFlowLight}\textbf{6.91} [6.75, 7.06] & \cellcolor{LoopedFlowLight}\textbf{5.84} [5.54, 6.07] & \cellcolor{LoopedFlowLight}\textbf{6.62} [6.54, 6.70] \\
\midrule
\multirow{2}{*}{Huginn 3.5B} & DAS-Wave & 4.23 [4.03, 4.43] & 5.38 [5.12, 5.55] & 5.45 [5.36, 5.53] & 4.57 [4.43, 4.72] & 5.20 [5.14, 5.26] \\
& \cellcolor{LoopedFlowLight}\ours & \cellcolor{LoopedFlowLight}\textbf{4.48} [4.20, 4.76] & \cellcolor{LoopedFlowLight}\textbf{6.21} [5.83, 6.47] & \cellcolor{LoopedFlowLight}\textbf{6.18} [6.06, 6.30] & \cellcolor{LoopedFlowLight}\textbf{5.06} [4.88, 5.25] & \cellcolor{LoopedFlowLight}\textbf{5.70} [5.62, 5.78] \\
\midrule
\multirow{2}{*}{Rec.\ TinyLlama} & LoopSpec & 5.79 [5.58, 6.01] & 6.92 [6.65, 7.16] & 7.35 [7.16, 7.50] & 6.48 [6.20, 6.74] & -- \\
& \cellcolor{LoopedFlowLight}\ours & \cellcolor{LoopedFlowLight}\textbf{6.73} [6.44, 7.01] & \cellcolor{LoopedFlowLight}\textbf{7.79} [7.52, 8.04] & \cellcolor{LoopedFlowLight}\textbf{8.20} [8.01, 8.37] & \cellcolor{LoopedFlowLight}\textbf{7.17} [6.84, 7.47] & \cellcolor{LoopedFlowLight}-- \\
\bottomrule
\end{tabular}
\end{table}
\subsection{Accuracy}
\label{sec:app_accuracy}
\Cref{tab:accuracy} reports the task accuracy of plain decoding and \ours across all models and benchmarks in \Cref{tab:main,tab:thinking}. On each task, the two methods differ on at most seven test problems due to bfloat16 rounding when verifying several tokens in parallel instead of one at a time. In our batched serving experiments (\Cref{sec:batch}), where we use batch-invariant kernels whose floating-point reductions are independent of batch size, \ours produces bit-identical outputs to plain decoding across all problems on every model.\begin{table}[!htb]
\centering
\caption{\textbf{\ours matches the accuracy of plain decoding within 3 points on every task.} Accuracy (\%) under greedy decoding. For each task, the left column is plain decoding and the shaded column is \ours. The two differ on at most seven problems per task, because bfloat16 rounding differs when several tokens are verified in parallel.}
\label{tab:accuracy}
\small
\setlength{\tabcolsep}{4.5pt}
\begin{tabular}{lcccccccccc}
\toprule
Model & \multicolumn{2}{c}{GSM8K} & \multicolumn{2}{c}{MATH-500} & \multicolumn{2}{c}{HumanEval+} & \multicolumn{2}{c}{MBPP+} & \multicolumn{2}{c}{BBH} \\
\cmidrule(lr){2-3} \cmidrule(lr){4-5} \cmidrule(lr){6-7} \cmidrule(lr){8-9} \cmidrule(lr){10-11}
\midrule
Ouro~1.4B & 75.0 & \cellcolor{LoopedFlowLight}74.0 & 76.0 & \cellcolor{LoopedFlowLight}73.0 & 64.6 & \cellcolor{LoopedFlowLight}64.0 & 65.1 & \cellcolor{LoopedFlowLight}65.1 & 70.9 & \cellcolor{LoopedFlowLight}71.1 \\
Ouro~2.6B & 85.0 & \cellcolor{LoopedFlowLight}85.0 & 57.0 & \cellcolor{LoopedFlowLight}59.0 & 74.4 & \cellcolor{LoopedFlowLight}74.4 & 69.3 & \cellcolor{LoopedFlowLight}69.0 & 78.9 & \cellcolor{LoopedFlowLight}80.2 \\
Recurrent~Llama~3.2 & 47.0 & \cellcolor{LoopedFlowLight}45.0 & 44.0 & \cellcolor{LoopedFlowLight}41.0 & 11.6 & \cellcolor{LoopedFlowLight}11.0 & 24.3 & \cellcolor{LoopedFlowLight}23.8 & 34.3 & \cellcolor{LoopedFlowLight}34.1 \\
Huginn 3.5B & 29.0 & \cellcolor{LoopedFlowLight}29.0 & 25.0 & \cellcolor{LoopedFlowLight}26.0 & 14.6 & \cellcolor{LoopedFlowLight}14.6 & 26.2 & \cellcolor{LoopedFlowLight}26.5 & 31.9 & \cellcolor{LoopedFlowLight}32.0 \\
Recurrent~TinyLlama & 38.0 & \cellcolor{LoopedFlowLight}40.0 & 41.0 & \cellcolor{LoopedFlowLight}42.0 & 14.0 & \cellcolor{LoopedFlowLight}13.4 & 23.3 & \cellcolor{LoopedFlowLight}23.0 & -- & \cellcolor{LoopedFlowLight}-- \\
\midrule
Ouro~1.4B~Thinking & 87.0 & \cellcolor{LoopedFlowLight}85.0 & 88.0 & \cellcolor{LoopedFlowLight}85.0 & -- & \cellcolor{LoopedFlowLight}-- & -- & \cellcolor{LoopedFlowLight}-- & -- & \cellcolor{LoopedFlowLight}-- \\
Ouro~2.6B~Thinking & 88.0 & \cellcolor{LoopedFlowLight}87.0 & 89.0 & \cellcolor{LoopedFlowLight}91.0 & -- & \cellcolor{LoopedFlowLight}-- & -- & \cellcolor{LoopedFlowLight}-- & -- & \cellcolor{LoopedFlowLight}-- \\
\bottomrule
\end{tabular}
\end{table}

\subsection{Comparison with the parallel sampler}
\label{sec:app_sampler}
The parallel sampler of \citet{geiping2026efficient} initializes a window of tokens and updates their hidden states in parallel across multiple recurrent steps, emitting each token once its hidden state changes by less than a threshold. Because the released code does not support the retrofitted models, we reimplemented it in PyTorch with four loops per step and a convergence threshold of $0.03$, and measure its speedup relative to plain decoding in the same implementation. Because the parallel sampler does not verify its predictions, it is not lossless (\Cref{tab:sampler_acc}). Under greedy decoding, its accuracy stays within $3$ points of plain decoding on Recurrent~Llama~3.2 and Recurrent~TinyLlama, but drops on all five tasks on Huginn 3.5B, including by $4$ points on GSM8K ($23.0\%$ vs.\ $27.0\%$) and $9$ points on MATH-500 ($16.0\%$ vs.\ $25.0\%$), where a third of the tokens reach $32$ loops before converging. Under sampling, accuracy drops further, by $13$ points on GSM8K and $12$ points on MBPP+ for Huginn 3.5B, and by $6$--$8$ points on MATH-500 and MBPP+ for Recurrent~Llama~3.2. \Cref{tab:sampler_speed} compares its speedups with those of \ours. Under greedy decoding, the parallel sampler is faster than \ours on GSM8K for Recurrent~Llama~3.2 and Huginn 3.5B and on MBPP+ for Huginn 3.5B, and slower in the other eleven settings, but it does not preserve the outputs of the model.

\begin{table}[!htb]
\centering
\caption{\textbf{The parallel sampler does not preserve accuracy, which drops by up to 13 points on Huginn 3.5B.} Accuracy (\%). For each task, the left column is plain decoding and the shaded column is the parallel sampler~\citep{geiping2026efficient}.}
\label{tab:sampler_acc}
\small
\setlength{\tabcolsep}{4.5pt}
\begin{tabular}{lcccccccccc}
\toprule
Model & \multicolumn{2}{c}{GSM8K} & \multicolumn{2}{c}{MATH-500} & \multicolumn{2}{c}{HumanEval+} & \multicolumn{2}{c}{MBPP+} & \multicolumn{2}{c}{BBH} \\
\cmidrule(lr){2-3} \cmidrule(lr){4-5} \cmidrule(lr){6-7} \cmidrule(lr){8-9} \cmidrule(lr){10-11}
\midrule
\multicolumn{11}{l}{\textit{Greedy decoding}} \\
Recurrent~Llama~3.2 & 46.0 & \cellcolor{black!8}47.0 & 41.0 & \cellcolor{black!8}39.0 & 11.0 & \cellcolor{black!8}8.5 & 23.0 & \cellcolor{black!8}23.0 & 37.0 & \cellcolor{black!8}38.5 \\
Huginn 3.5B & 27.0 & \cellcolor{black!8}23.0 & 25.0 & \cellcolor{black!8}16.0 & 14.6 & \cellcolor{black!8}12.2 & 27.0 & \cellcolor{black!8}25.4 & 31.9 & \cellcolor{black!8}31.1 \\
Recurrent~TinyLlama & 37.0 & \cellcolor{black!8}40.0 & 40.0 & \cellcolor{black!8}39.0 & 14.0 & \cellcolor{black!8}14.0 & 24.1 & \cellcolor{black!8}24.1 & -- & \cellcolor{black!8}-- \\
\midrule
\multicolumn{11}{l}{\textit{Sampling at temperature $1$ and top-$p$ $0.7$}} \\
Recurrent~Llama~3.2 & 52.0 & \cellcolor{black!8}50.0 & 39.0 & \cellcolor{black!8}31.0 & 13.4 & \cellcolor{black!8}11.6 & 16.9 & \cellcolor{black!8}11.1 & 38.5 & \cellcolor{black!8}36.3 \\
Huginn 3.5B & 19.0 & \cellcolor{black!8}6.0 & 18.0 & \cellcolor{black!8}10.0 & 7.3 & \cellcolor{black!8}6.1 & 22.0 & \cellcolor{black!8}10.3 & 30.4 & \cellcolor{black!8}31.1 \\
Recurrent~TinyLlama & 39.0 & \cellcolor{black!8}34.0 & 34.0 & \cellcolor{black!8}33.0 & 7.9 & \cellcolor{black!8}7.9 & 16.9 & \cellcolor{black!8}16.1 & -- & \cellcolor{black!8}-- \\
\bottomrule
\end{tabular}
\end{table}

\begin{table}[!htb]
\centering
\caption{\textbf{Under greedy decoding, \ours is faster than the parallel sampler in 11 of 14 settings, without changing the outputs.} Wall-clock speedup ($\times$) over plain decoding on an A100 at batch size one, with accepted tokens per step $\tauacc$ in parentheses. Sampling uses temperature $1.0$ and top-$p$ $0.7$. $\dagger$ marks the parallel sampler~\citep{geiping2026efficient}, which is not lossless. It is faster under sampling, where its accuracy drops the most (\Cref{tab:sampler_acc}).}
\label{tab:sampler_speed}
\fitwidth{%
\setlength{\tabcolsep}{2.2pt}
\begin{tabular}{llcccccccccc}
\toprule
 & & \multicolumn{2}{c}{GSM8K} & \multicolumn{2}{c}{MATH-500} & \multicolumn{2}{c}{HumanEval+} & \multicolumn{2}{c}{MBPP+} & \multicolumn{2}{c}{BBH} \\
\cmidrule(lr){3-4}\cmidrule(lr){5-6}\cmidrule(lr){7-8}\cmidrule(lr){9-10}\cmidrule(lr){11-12}
Model & Method & Greedy & Sampling & Greedy & Sampling & Greedy & Sampling & Greedy & Sampling & Greedy & Sampling \\
\midrule
\multirow{2}{*}{\makecell[l]{Recurrent\\Llama 3.2}} & Parallel sampler$^\dagger$ & 5.38 & 5.03 & 5.68 & 5.41 & 5.67 & 5.56 & 5.54 & 5.52 & 5.44 & 5.39 \\
 & \cellcolor{LoopedFlowLight}\ours & \cellcolor{LoopedFlowLight}5.04 (20.4) & \cellcolor{LoopedFlowLight}3.28 (7.6) & \cellcolor{LoopedFlowLight}6.97 (28.5) & \cellcolor{LoopedFlowLight}4.66 (11.0) & \cellcolor{LoopedFlowLight}6.91 (28.0) & \cellcolor{LoopedFlowLight}4.69 (10.9) & \cellcolor{LoopedFlowLight}5.84 (23.5) & \cellcolor{LoopedFlowLight}4.34 (10.1) & \cellcolor{LoopedFlowLight}6.62 (27.1) & \cellcolor{LoopedFlowLight}5.13 (12.1) \\
\midrule
\multirow{2}{*}{\makecell[l]{Huginn\\3.5B}} & Parallel sampler$^\dagger$ & 5.37 & 5.38 & 5.79 & 5.79 & 5.78 & 5.73 & 5.24 & 5.38 & 5.34 & 5.13 \\
 & \cellcolor{LoopedFlowLight}\ours & \cellcolor{LoopedFlowLight}4.48 (19.0) & \cellcolor{LoopedFlowLight}3.16 (7.3) & \cellcolor{LoopedFlowLight}6.21 (26.6) & \cellcolor{LoopedFlowLight}4.60 (10.8) & \cellcolor{LoopedFlowLight}6.18 (26.2) & \cellcolor{LoopedFlowLight}4.70 (10.9) & \cellcolor{LoopedFlowLight}5.06 (21.4) & \cellcolor{LoopedFlowLight}3.52 (8.1) & \cellcolor{LoopedFlowLight}5.70 (24.6) & \cellcolor{LoopedFlowLight}4.48 (19.3) \\
\midrule
\multirow{2}{*}{\makecell[l]{Recurrent\\TinyLlama}} & Parallel sampler$^\dagger$ & 5.72 & 5.32 & 6.01 & 5.87 & 6.03 & 5.06 & 5.86 & 5.54 & \multicolumn{2}{c}{\multirow{2}{*}{\makecell{prompts exceed\\max context length}}} \\
 & \cellcolor{LoopedFlowLight}\ours & \cellcolor{LoopedFlowLight}6.73 (22.2) & \cellcolor{LoopedFlowLight}4.01 (8.1) & \cellcolor{LoopedFlowLight}7.79 (26.0) & \cellcolor{LoopedFlowLight}4.83 (9.9) & \cellcolor{LoopedFlowLight}8.20 (27.0) & \cellcolor{LoopedFlowLight}5.61 (11.4) & \cellcolor{LoopedFlowLight}7.17 (23.5) & \cellcolor{LoopedFlowLight}5.25 (10.6) & \multicolumn{2}{c}{\cellcolor{LoopedFlowLight}} \\
\bottomrule
\end{tabular}}
\end{table}

\subsection{Multi-token prediction with early layer states}
\label{sec:app_probe}
\begin{figure}[!htb]
\centering
\includegraphics[width=\linewidth]{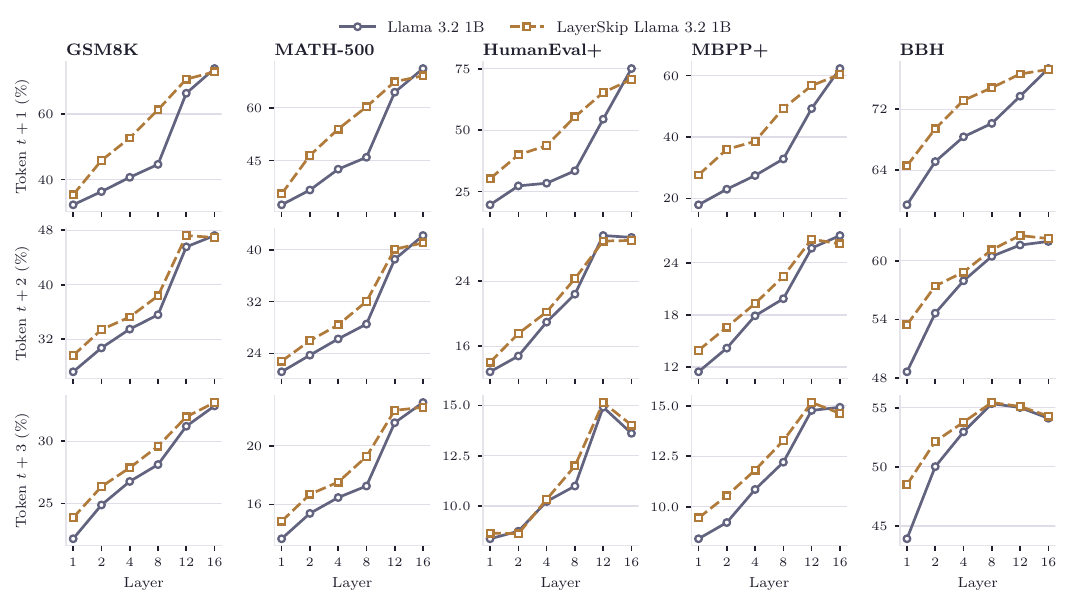}
\caption{\textbf{Early-exit supervision makes the early layers more predictive of the next tokens.} Accuracy of linear readers that predict the token one, two, or three positions ahead from the hidden state after each layer of Llama 3.2 1B and LayerSkip Llama 3.2 1B.}
\label{fig:motivation_full}
\end{figure}

In \Cref{sec:intro,sec:dense}, we argue that supervising next-token prediction at early depths encourages shallow states to carry information about several upcoming tokens. To test this on dense models, we compare Llama 3.2 1B against LayerSkip Llama 3.2 1B~\citep{elhoushi2024layerskip}, which fine-tunes Llama 3.2 1B with next-token losses at shallow layers, and measure how well the hidden state after each layer predicts the next three tokens.\paragraph{Linear readers.}
Following Future Lens~\citep{pal2023future}, we train a separate linear reader for each layer and each lookahead offset $j\in\{1,2,3\}$. The reader applies a learned linear map to the hidden state of token ${\bf x}^t$ after a given layer, followed by the frozen final normalization and output head of the model, to predict ${\bf x}^{t+j}$. Because the output head is frozen, the reader only extracts information already present in the hidden state. We initialize each reader to the identity matrix, which recovers the model's next-token prediction at the final layer, and train it with cross-entropy for four epochs, selecting the checkpoint with the highest accuracy on a held-out $10\%$ validation split.\paragraph{Data.}
Because the two models share a tokenizer, we train and evaluate both on the same text. The training set consists of 6,300 sequences from the benchmark training splits (reference solutions of GSM8K and MATH problems, MBPP programs outside MBPP+, and BBH problems outside our evaluation split). The test set consists of reference solutions for 500 GSM8K test problems and MATH-500, canonical programs for HumanEval+ and MBPP+, and the evaluation problems of BBH.\paragraph{Results.}
\Cref{fig:motivation_full} shows the reader accuracy across layers on each benchmark. On all five benchmarks, LayerSkip Llama 3.2 1B predicts the immediate next token ($j=1$) more accurately than Llama 3.2 1B from every intermediate layer by up to $22$ points, while Llama 3.2 1B is equally or more accurate at the final layer. This gap carries over to two and three tokens ahead ($j=2$ and $j=3$), though it narrows with distance. The only exception is $j=3$ on HumanEval+, where both models remain within one point across all layers. These results support the hypothesis in \Cref{sec:intro,sec:dense} that next-token supervision at early depths causes intermediate states to encode information about future tokens.
\subsection{Acceptance rate by draft position}
\label{sec:app_positions}
\begin{figure}[!htb]
\centering
\includegraphics[width=\linewidth]{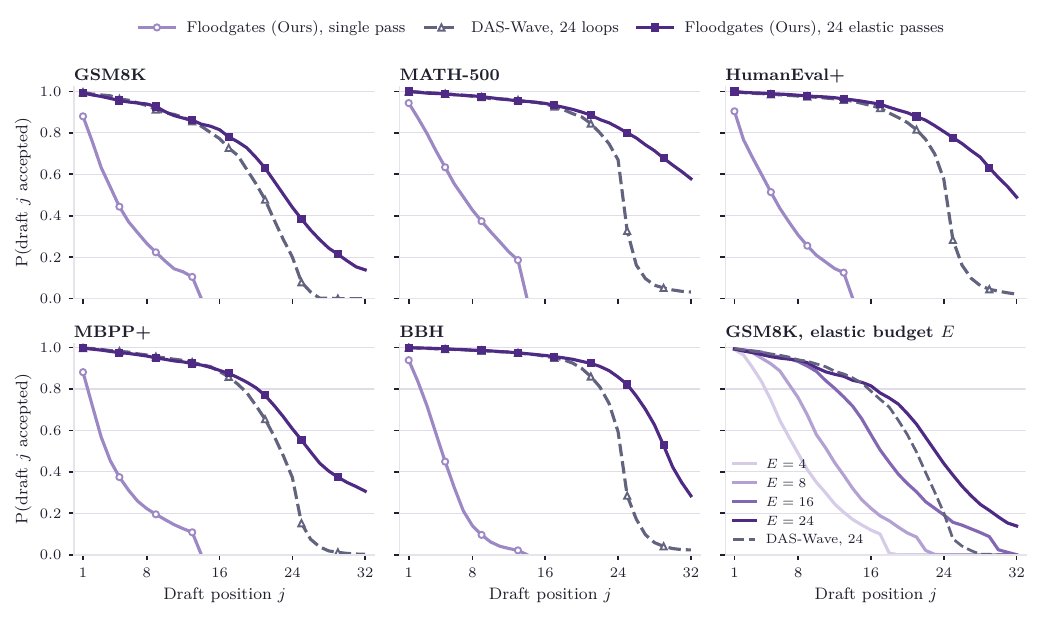}
\caption{\textbf{With 24 elastic passes, the acceptance rate stays above 50\% up to draft positions 23 to 32.} Acceptance rate by draft position on Recurrent~Llama~3.2 ($R=32$), measured as the fraction of verification steps in which the $j$-th draft is accepted. DAS-Wave runs the same $24$ loops starting from copies of the last verified token. The last panel varies the elastic drafting budget $E$ on GSM8K.}

\label{fig:positions}
\end{figure}

To examine how elastic drafting increases accepted draft length, \Cref{fig:positions} plots the acceptance rate at each draft position $j$ on Recurrent~Llama~3.2 ($R=32$). Single-pass drafting ($E=0$) is accurate on early slots, with $88$--$94\%$ acceptance at the first position, but its acceptance rate drops after position $12$. Elastic drafting with $E=24$ passes extends high acceptance across the window, keeping the acceptance rate above $50\%$ up to positions $23$--$32$ depending on the task. DAS-Wave runs the same loops initialized from copies of the last verified token; it matches \ours over the first $16$ positions, but falls behind on later positions and drops sharply after position $26$ on MATH-500, HumanEval+, and BBH. Because \ours initializes the full window from its trained parallel draft, it retains $44$--$86\%$ acceptance at position $24$, compared to $20$--$67\%$ for DAS-Wave. The last panel of \Cref{fig:positions} varies the elastic drafting budget $E$ on GSM8K, showing that additional passes extend the $50\%$ acceptance horizon from position $7$ at $E=4$ to position $23$ at $E=24$.

\FloatBarrier

\subsection{System throughput across batch sizes}
\label{sec:app_serving_results}
\Cref{tab:serving_all} reports the per-batch-size system throughput values underlying \Cref{fig:serving}, along with the maximum batch size that fits in memory and the peak throughput of each method relative to plain decoding. We also evaluate \ours with a fixed verification block, which verifies all $k$ drafts for every sequence at all batch sizes instead of using the adaptive budget $M^\star$ in (\ref{eq:batch_select}). Fixed-block verification matches $M^\star$ at small batch sizes, but falls behind as the batch size grows, reducing throughput at the largest batch size by $19\%$ on Ouro~1.4B and $54\%$ on Recurrent~Llama~3.2.\begin{table}[!tp]
\centering
\caption{\textbf{\ours reaches the highest peak throughput on three of the four models, and the adaptive budget adds up to 54\% at the largest batch size.} System throughput (tokens/s) across batch sizes. The last column reports the speedup of each method's peak throughput over that of plain decoding. ``Fixed block'' verifies all $k$ drafts of every request instead of the budget $M^\star$ in (\ref{eq:batch_select}). LoopSpec ($\ast$) runs as many server instances as fit in memory.}
\label{tab:serving_all}
\small
\setlength{\tabcolsep}{4pt}
\begin{tabular}{lrrrrrrrr}
\toprule
Batch size & 1 & 2 & 4 & 8 & 16 & 32 & Largest & Best \\
\midrule
\multicolumn{9}{l}{\textit{Ouro~1.4B, GSM8K}, largest batch size 80} \\
Plain decoding & 105 & 207 & 394 & 733 & 1{,}252 & 1{,}948 & 2{,}708 & $1.00\times$ \\
LoopSpec & 229 & 322 & 384 & 414 & -- & -- & 419$^{\ast}$ & $0.15\times$ \\
DAS & 193 & 371 & 674 & 1{,}183 & 1{,}832 & 2{,}502 & 3{,}302 & $1.22\times$ \\
Prompt lookup decoding & 206 & 387 & 662 & 1{,}045 & 1{,}460 & 1{,}775 & 2{,}079 & $0.77\times$ \\
\ours, fixed block & 588 & 1{,}099 & 1{,}865 & 2{,}980 & 3{,}787 & 4{,}463 & 4{,}859 & $1.81\times$ \\
\rowcolor{LoopedFlowLight} \ours & \textbf{593} & \textbf{1{,}111} & \textbf{1{,}913} & \textbf{3{,}037} & \textbf{4{,}185} & \textbf{5{,}559} & \textbf{5{,}805} & \textbf{$2.14\times$} \\
\midrule
\multicolumn{9}{l}{\textit{Ouro~2.6B, GSM8K}, largest batch size 40} \\
Plain decoding & 53 & 104 & 199 & 370 & 629 & 985 & 1{,}051 & $1.00\times$ \\
LoopSpec & 130 & 177 & 209 & -- & -- & -- & 220$^{\ast}$ & $0.21\times$ \\
DAS & 114 & 219 & 399 & 699 & 1{,}079 & 1{,}458 & 1{,}628 & $1.55\times$ \\
Prompt lookup decoding & 102 & 192 & 331 & 523 & 725 & 886 & 950 & $0.90\times$ \\
\ours, fixed block & 323 & 603 & 1{,}023 & 1{,}636 & 2{,}073 & 2{,}456 & 2{,}218 & $2.34\times$ \\
\rowcolor{LoopedFlowLight} \ours & \textbf{327} & \textbf{613} & \textbf{1{,}055} & \textbf{1{,}673} & \textbf{2{,}283} & \textbf{2{,}966} & \textbf{2{,}827} & \textbf{$2.82\times$} \\
\midrule
\multicolumn{9}{l}{\textit{Recurrent~Llama~3.2, GSM8K}, largest batch size 128} \\
Plain decoding & 39 & 78 & 152 & 290 & 522 & 850 & 1{,}580 & $1.00\times$ \\
LoopSpec & 180 & 264 & 314 & 320 & -- & -- & 315$^{\ast}$ & $0.20\times$ \\
DAS & 152 & 291 & 548 & 937 & 1{,}431 & 1{,}843 & \textbf{2{,}767} & \textbf{$1.75\times$} \\
DAS-Wave & 178 & 308 & 425 & 587 & 704 & 822 & 866 & $0.56\times$ \\
Prompt lookup decoding & 73 & 138 & 256 & 445 & 671 & 824 & 1{,}033 & $0.65\times$ \\
\ours, fixed block & 190 & 354 & 631 & 1{,}009 & 1{,}271 & 1{,}426 & 1{,}697 & $1.07\times$ \\
\rowcolor{LoopedFlowLight} \ours & \textbf{194} & \textbf{365} & \textbf{658} & \textbf{1{,}143} & \textbf{1{,}778} & \textbf{2{,}176} & 2{,}609 & $1.65\times$ \\
\midrule
\multicolumn{9}{l}{\textit{Ouro~1.4B~Thinking, MATH-500}, largest batch size 12} \\
Plain decoding & 97 & 175 & 290 & 439 & -- & -- & 530 & $1.00\times$ \\
LoopSpec & 174 & 232 & 272 & 292 & -- & -- & 292$^{\ast}$ & $0.55\times$ \\
DAS & 181 & 321 & 519 & 760 & -- & -- & 890 & $1.68\times$ \\
Prompt lookup decoding & 170 & 302 & 471 & 642 & -- & -- & 764 & $1.44\times$ \\
\ours, fixed block & \textbf{369} & 636 & 999 & 1{,}406 & -- & -- & 1{,}598 & $3.02\times$ \\
\rowcolor{LoopedFlowLight} \ours & 367 & \textbf{641} & \textbf{1{,}054} & \textbf{1{,}443} & -- & -- & \textbf{1{,}705} & \textbf{$3.22\times$} \\
\bottomrule
\end{tabular}
\end{table}
\FloatBarrier

\subsection{Serving with a single shared cache across loops}
\label{sec:app_share}
\begin{figure}[!htb]
\centering
\includegraphics[width=\linewidth]{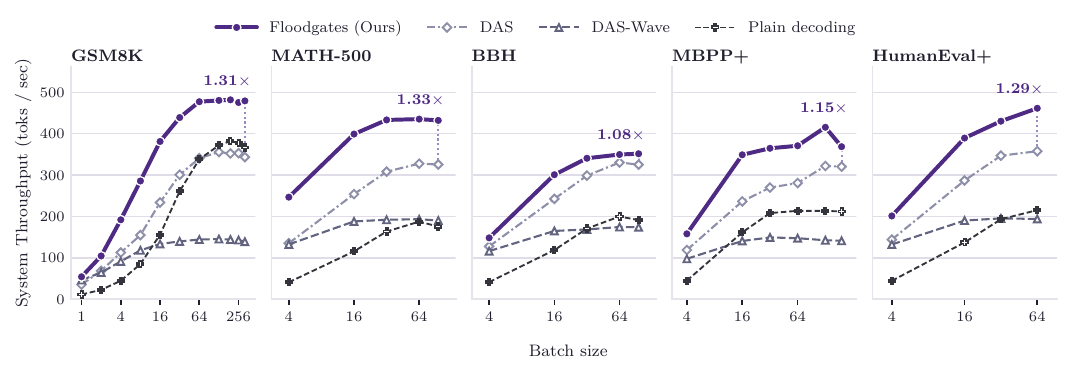}
\caption{\textbf{With a single shared cache, \ours has the highest throughput at every batch size on all five tasks.} System throughput against batch size on an A100 for Huginn 3.5B, served with one key-value cache for all loops. The number in each panel is the throughput of \ours relative to the strongest baseline at the largest batch size.}
\label{fig:serving_huginn}
\end{figure}

A looped model stores the keys and values of every loop, which limits how many requests fit in memory. \citet{prairie2026benice} show that some looped models can instead keep only the final-loop key-value cache and reuse it at every loop without further training. Unlike the deep prefix cache in the main text, which only the drafting pass reads, this setting reuses the final-loop cache across all $R$ verification loops as well. Because LoopSpec and WaveFront start the next token before the current token finishes its loops, they require the cache of every loop and cannot use a single cache. We therefore compare \ours with DAS, DAS-Wave, and plain decoding.\paragraph{Accuracy.}
\Cref{tab:share_acc} compares per-loop caches with a single shared cache at batch size one on the three 32-loop models. Using a single cache changes the accuracy of plain decoding by at most $2$ points on Huginn 3.5B and Recurrent~TinyLlama, and lowers it by $3$--$5$ points on GSM8K and MATH-500 for Recurrent~Llama~3.2. With the single cache, \ours matches plain decoding within $1$ point on almost every task.\begin{table}[!htb]
\centering
\caption{\textbf{A single shared cache changes the accuracy of plain decoding by at most 2 points on Huginn 3.5B and Recurrent~TinyLlama.} Accuracy (\%) at batch size one, with a separate cache per loop or with the final-loop cache reused by all loops. Both cache settings are evaluated on the same problems, with the number of problems in parentheses.}
\label{tab:share_acc}
\small
\setlength{\tabcolsep}{2.5pt}
\begin{tabular}{llccccc}
\toprule
Model & Decoding & GSM8K & MATH-500 & HumanEval+ & MBPP+ & BBH \\
& & (100) & (100) & (164) & (378) & (540) \\
\midrule
\multirow{3}{*}{Recurrent~Llama~3.2} & Plain decoding, per-loop caches & 47.0 & 44.0 & 11.6 & 24.3 & 34.3 \\
& Plain decoding, single cache & 44.0 & 39.0 & 9.8 & 24.6 & 35.2 \\
& \cellcolor{LoopedFlowLight}\ours, single cache & \cellcolor{LoopedFlowLight}44.0 & \cellcolor{LoopedFlowLight}42.0 & \cellcolor{LoopedFlowLight}10.4 & \cellcolor{LoopedFlowLight}24.3 & \cellcolor{LoopedFlowLight}35.0 \\
\midrule
\multirow{3}{*}{Recurrent~TinyLlama} & Plain decoding, per-loop caches & 38.0 & 41.0 & 14.0 & 23.3 & -- \\
& Plain decoding, single cache & 38.0 & 42.0 & 13.4 & 24.3 & -- \\
& \cellcolor{LoopedFlowLight}\ours, single cache & \cellcolor{LoopedFlowLight}39.0 & \cellcolor{LoopedFlowLight}42.0 & \cellcolor{LoopedFlowLight}14.0 & \cellcolor{LoopedFlowLight}23.8 & \cellcolor{LoopedFlowLight}-- \\
\midrule
\multirow{3}{*}{Huginn 3.5B} & Plain decoding, per-loop caches & 29.0 & 25.0 & 14.6 & 26.2 & 31.9 \\
& Plain decoding, single cache & 27.0 & 23.0 & 13.4 & 26.7 & 32.6 \\
& \cellcolor{LoopedFlowLight}\ours, single cache & \cellcolor{LoopedFlowLight}28.0 & \cellcolor{LoopedFlowLight}23.0 & \cellcolor{LoopedFlowLight}14.0 & \cellcolor{LoopedFlowLight}27.2 & \cellcolor{LoopedFlowLight}32.2 \\
\bottomrule
\end{tabular}
\end{table}\paragraph{Throughput.}
Since a single cache preserves the accuracy of Huginn 3.5B, we serve it with continuous batching as in \Cref{sec:batch} on all problems of the five tasks (\Cref{fig:serving_huginn}). With per-loop caches, \ours runs out of memory on GSM8K at batch size $24$ and peaks at $380$ tokens/s at batch size $16$. With a single cache, it scales to batch size $320$ on GSM8K (reaching $481$ tokens/s), $192$ on MBPP+, $96$ on MATH-500 and BBH, and $64$ on HumanEval+. \ours achieves the highest system throughput at every batch size on all five tasks. At the largest batch size, it outperforms the strongest baseline by $1.29$--$1.33\times$ on HumanEval+, GSM8K, and MATH-500, $1.15\times$ on MBPP+, and $1.08\times$ on BBH. Together, these results show that \ours combines naturally with cache sharing to serve much larger batches while maintaining the highest throughput.
\section{Design choices}
\label{sec:app_design}
We ablate the design choices of \ours on Ouro~1.4B in \Cref{tab:ablation}. Our default configuration samples $64$ training positions per sequence and uses $k=12$ mask tokens with rank-$64$ LoRA adapters. Each row of \Cref{tab:ablation} modifies a single choice from this default, with all drafters trained for $8$K steps on the four-domain dataset of \Cref{sec:app_training}.\paragraph{Markov head.}
In (\ref{eq:draft}), each slot state $\hat{\bf z}^{t+j}$ is read out independently by the head $g$, so the draft token ${\bf d}^j$ does not depend directly on the preceding draft token ${\bf d}^{j-1}$. Following DSpark~\citep{cheng2026dspark}, we also evaluate a Markov head that adds a logit bias conditioned on the previous draft token
\begin{equation}
\label{eq:markov}
{\bf q}^j(v) \propto g(\hat{\bf z}^{t+j})(v)\,\exp\bigl(({\bf a}_{{\bf d}^{j-1}}^{\top}{\bf B})_v\bigr), \qquad {\bf d}^j=\argmax({\bf q}^j), \qquad {\bf d}^0={\bf x}^t,
\end{equation}
where ${\bf a}_u\in\mathbb{R}^{128}$ is a learned token embedding and ${\bf B}\in\mathbb{R}^{128\times|V|}$ projects it to a vocabulary bias. We initialize ${\bf B}$ to zero so that training starts from the standard drafter. During training, ${\bf d}^{j-1}$ is replaced by the ground-truth token ${\bf x}^{t+j-1}$. During drafting, tokens are selected from left to right after the parallel pass of $f_\theta$, adding one small matrix-vector product per slot.
\paragraph{Results.}
\Cref{tab:ablation} shows the ablation results. Sampling fewer positions per sequence ($16$), using fewer mask tokens ($k=8$), halving the adapter rank ($32$), and replacing the learned mask embedding with random noise all reduce both accepted tokens per step and decoding throughput. Increasing the number of mask tokens to $k=16$ raises accepted tokens per step on GSM8K from $7.69$ to $8.26$, and adding the Markov head improves acceptance on both tasks (from $7.69$ to $8.00$ on GSM8K and from $6.46$ to $6.98$ on MATH-500), yielding the highest throughput ($482$ tokens/s).\begin{table}[!htb]
\centering
\caption{\textbf{The default drafter is within 2\% of the fastest configuration.} Accepted tokens per step on GSM8K and MATH-500, and decoding throughput (tokens/s) on GSM8K, for drafters of Ouro~1.4B trained for $8$K steps. Each row changes one choice of the default configuration in the first row.}
\label{tab:ablation}
\small
\setlength{\tabcolsep}{5pt}
\begin{tabular}{lccc}
\toprule
Configuration & GSM8K & MATH-500 & tok/s \\
\midrule
\rowcolor{LoopedFlowLight} \ours, $64$ positions, $k=12$, rank $64$ & 7.69 & 6.46 & 475 \\
$16$ sampled positions per sequence & 7.30 & 5.90 & 449 \\
$k=8$ & 6.58 & 5.84 & 410 \\
$k=16$ & \textbf{8.26} & 6.63 & 477 \\
rank $32$ & 7.50 & 6.23 & 465 \\
noise input instead of mask tokens & 7.54 & 6.21 & 451 \\
Markov head & 8.00 & \textbf{6.98} & \textbf{482} \\
\bottomrule
\end{tabular}
\end{table}

\end{document}